\documentclass[10pt,twocolumn,letterpaper]{article}
\PassOptionsToPackage{dvipsnames,table}{xcolor}
\usepackage[pagenumbers]{iccv} 

\definecolor{iccvblue}{rgb}{0.21,0.49,0.74}
\usepackage[accsupp]{axessibility}  
\usepackage[pagebackref,breaklinks,colorlinks,allcolors=iccvblue]{hyperref}

\def\paperID{14339} 
\def\confName{ICCV}
\def\confYear{2025}

\title{CoopTrack: Exploring End-to-End Learning for Efficient Cooperative Sequential Perception}

\author{
Jiaru Zhong\textsuperscript{1,2} \hspace{0.5em} Jiahao Wang\textsuperscript{4} \hspace{0.5em} Jiahui Xu\textsuperscript{3} \hspace{0.5em}  Xiaofan Li\textsuperscript{5} \hspace{0.5em}  Zaiqing Nie\textsuperscript{1}\equalcorres \hspace{0.5em} Haibao Yu\textsuperscript{3,1}\equalcorres \\
\textsuperscript{1} Institute for AI Industry Research, Tsinghua University \textsuperscript{2} The Hong Kong Polytechnic University \\ \textsuperscript{3} The University of Hong Kong \textsuperscript{4} School of Vehicle and Mobility, Tsinghua University \textsuperscript{5} Baidu Inc. \\
\tt\small{zhong.jiaru@outlook.com, zaiqing@air.tsinghua.edu.cn, yuhaibao94@gmail.com}
}

\begin{document}
\maketitle
\begin{abstract}
Cooperative perception aims to address the inherent limitations of single-vehicle autonomous driving systems through information exchange among multiple agents. Previous research has primarily focused on single-frame perception tasks. However, the more challenging cooperative sequential perception tasks, such as cooperative 3D multi-object tracking, have not been thoroughly investigated.
Therefore, we propose CoopTrack, a fully instance-level end-to-end framework for cooperative tracking, featuring learnable instance association, which fundamentally differs from existing approaches.
CoopTrack transmits sparse instance-level features that significantly enhance perception capabilities while maintaining low transmission costs. Furthermore, the framework comprises two key components: Multi-Dimensional Feature Extraction, and Cross-Agent Association and Aggregation, which collectively enable comprehensive instance representation with semantic and motion features, and adaptive cross-agent association and fusion based on a feature graph.
Experiments on both the V2X-Seq and Griffin datasets demonstrate that CoopTrack achieves excellent performance. Specifically, it attains state-of-the-art results on V2X-Seq, with 39.0\% mAP and 32.8\% AMOTA. The project is available at \href{https://github.com/zhongjiaru/CoopTrack}{https://github.com/zhongjiaru/CoopTrack}.
\end{abstract}    
\section{Introduction}
\label{sec:introduction}
\begin{figure}[t]
  \centering
   \includegraphics[width=0.95\linewidth]{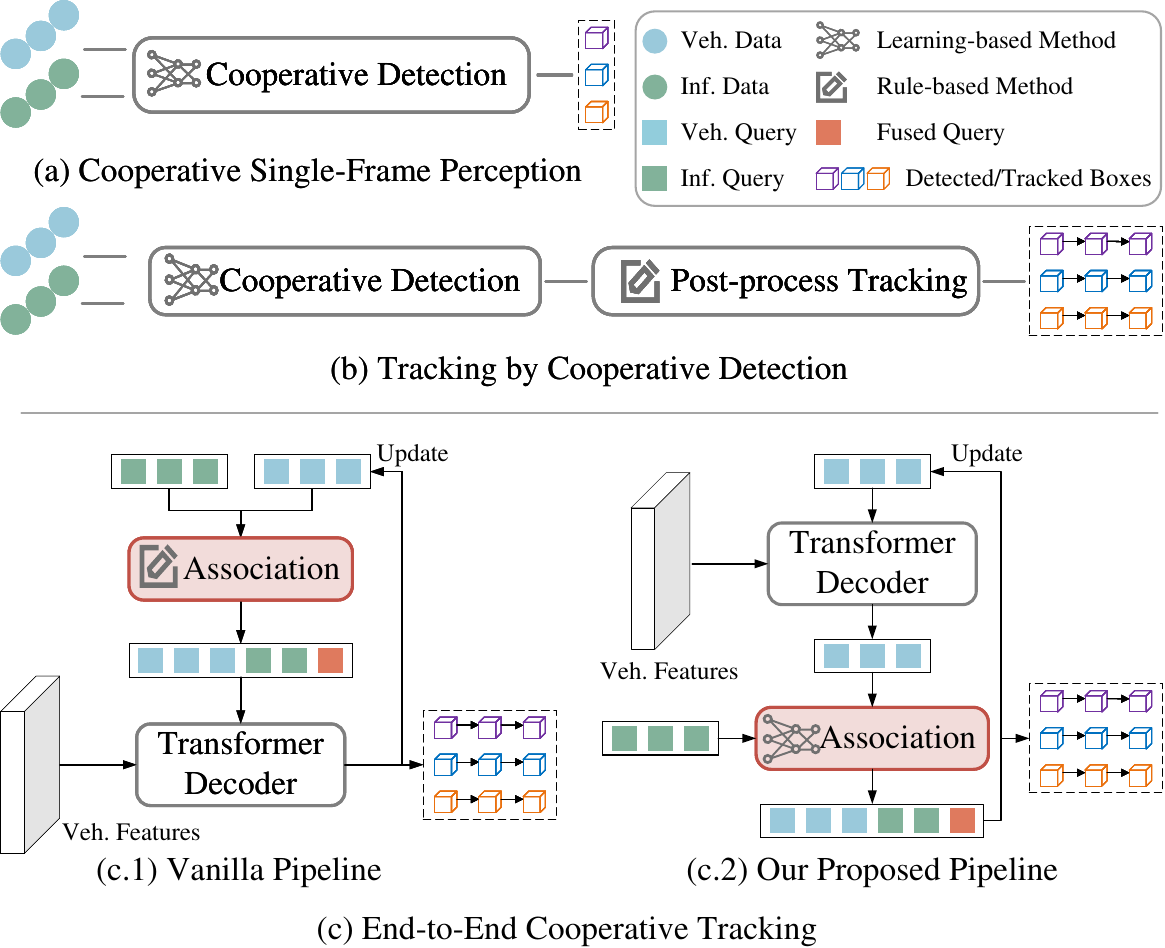}
   \vspace{-6pt}
   \caption{\textbf{Comparison with Current Cooperative Perception.} 
   (a) Most research focuses on single-frame perception tasks. 
   (b) Non end-to-end solutions adopt a tracking-by-cooperative-detection approach. 
   (c) End-to-end cooperative tracking methods enable joint detection and tracking:
   (c.1) A previous method \cite{yu2024end} applies rule-based query association.
   (c.2) Our proposed CoopTrack introduces a fully end-to-end framework with learnable instance association and a novel fusion-after-decoding pipeline.}
   \label{fig:fig1}
   \vspace{-10pt}
\end{figure}

Accurate and comprehensive perception of the surrounding environment is a fundamental task for autonomous driving. Despite numerous efforts having been made \cite{liu2025bevmamba, zhu2024lanemapnet,hao2025styledrive}, single-vehicle perception still faces significant challenges due to the limited information from a single observation perspective. With the assistance of vehicle-to-everything (V2X) communication, cooperative perception offers a promising solution and has thus garnered widespread attention. Currently, as illustrated in \cref{fig:fig1}(a), cooperative single-frame perception tasks, such as  3D object detection \cite{hu2022where2comm, xu2022v2x,li2022v2x}, have seen significant advancements. However, cooperative sequential perception tasks, such as cooperative 3D multi-object tracking (3D MOT), remain underexplored.

\begin{figure}[t]
  \centering
  \includegraphics[width=0.95\linewidth]{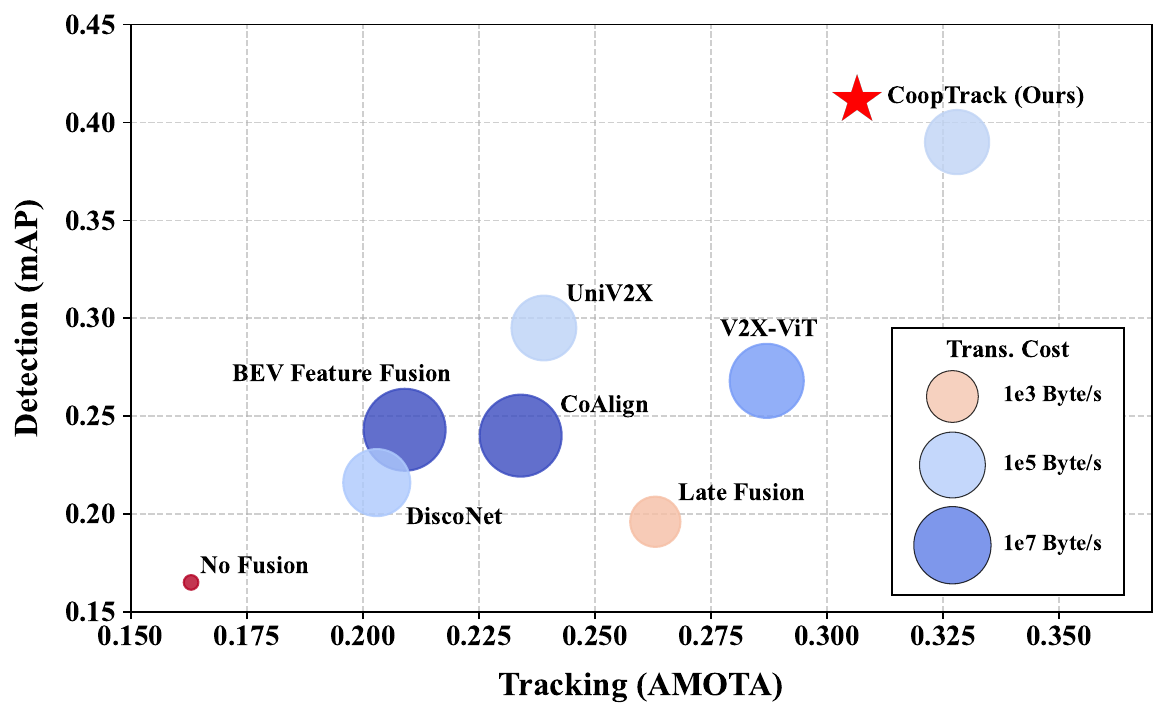}
  \vspace{-8pt}
   \caption{\textbf{Comparison with Cooperative Perception Methods} on V2X-Seq \cite{yu2023v2x} dataset. The X-axis and Y-axis represent detection and tracking performance, respectively, while the bubble size and color encode the transmission cost on a logarithmic scale. }
   \label{fig:fig2}
   \vspace{-15pt}
\end{figure}

The cooperative sequential perception problem can be conceptualized as the fusion of multi-agent perceptual information under constrained communication conditions, aiming to enhance perception over a continuous time horizon while minimizing transmitted data. Additionally, the challenge lies in seamlessly integrating the cooperative module into the sequential perception framework to establish an efficient pipeline.
As shown in \cref{fig:fig1}(b), traditional approaches \cite{xu2023v2v4real, yu2023v2x, chiu2024probabilistic, su2024collaborative, su20233d, su2024cooperative} follow the tracking-by-cooperative-detection pipeline, where cooperative detection results are fed into model-based trackers \cite{weng20203d} for post-processing. However, this paradigm fails to fully leverage fusion information for tracking.  In contrast, recent studies \cite{pang2023standing, ding2024ada,wang2024onetrack} demonstrate that end-to-end learning, utilizing queries to continuously track objects frame by frame, can achieve state-of-the-art (SOTA) tracking performance.

Recently, UniV2X \cite{yu2024end} pioneered an end-to-end framework for planning while also explicitly addressing cooperative tracking. As illustrated in \cref{fig:fig1}(c.1), it introduces an interpretable association and fusion module that merges queries from two agents, and the final queries are used to detect or track objects from the ego features.
As shown in \cref{fig:fig2}, although UniV2X represents a commendable exploratory effort, its rule-based association and fusion-before-decoding pipeline result in suboptimal performance.

Therefore, we propose a fully instance-level end-to-end framework for cooperative 3D MOT, termed CoopTrack. CoopTrack features learnable association and fusion, significantly enhancing perception capabilities while maintaining low transmission costs.
Specifically, as shown in \cref{fig:fig1}(c.2), CoopTrack employs a novel fusion-after-decoding pipeline that first performs decoding, followed by association and fusion, enabling seamless integration of cooperation and tracking.
To obtain adaptive association and fusion of instance-level features, CoopTrack comprises two pivotal modules. Firstly, the Multi-Dimensional Feature Extraction (MDFE) module is developed to capture semantic and motion features, thereby providing a comprehensive description of instances. Secondly, the learnable Cross-Agent Association and Aggregation comprises two key steps: Cross-Agent Alignment (CAA) and Graph-Based Association (GBA).
The CAA module bridges the domain gap between features of different agents through latent space transformation. The GBA module learns inter-agent feature similarities via graph attention mechanisms, generating matched pairs to guide the final feature aggregation. Through multi-dimensional features and feature alignment, the distinctiveness of different instances and the cross-agent consistency of the same instance are ensured. Based on them, the graph-based module introduces the adaptive association into the end-to-end framework, thereby enhancing performance.

To validate the efficacy of our method, we conduct extensive comparative and ablation studies on the real-world dataset V2X-Seq \cite{yu2023v2x}. As shown in \cref{fig:fig2}, CoopTrack significantly outperforms existing methods, including UniV2X \cite{yu2024end}, achieving a 9.5\% higher mAP and a 4.1\% improvement in AMOTA, while requiring substantially lower transmission costs. Experiments on the Griffin \cite{wang2025griffin} dataset demonstrate that CoopTrack also achieves strong performance in aerial-ground cooperative scenarios.

In summary, the contributions are as follows:
\begin{itemize}
    \item We propose CoopTrack, an instance-level end-to-end framework for cooperative sequential perception, featuring learnable instance association and a novel fusion-after-decoding pipeline.
    \item We introduce a learnable graph-based instance association module and an instance-level feature extraction module, capturing semantic and motion features to enhance cross-agent association learning.
    \item Our method surpasses all existing cooperative perception approaches, setting a new SOTA on the V2X-Seq dataset.
\end{itemize}
\section{Related work}
\label{sec:related work}

\begin{figure*}[t]
  \centering
  \includegraphics[width=0.98\linewidth]{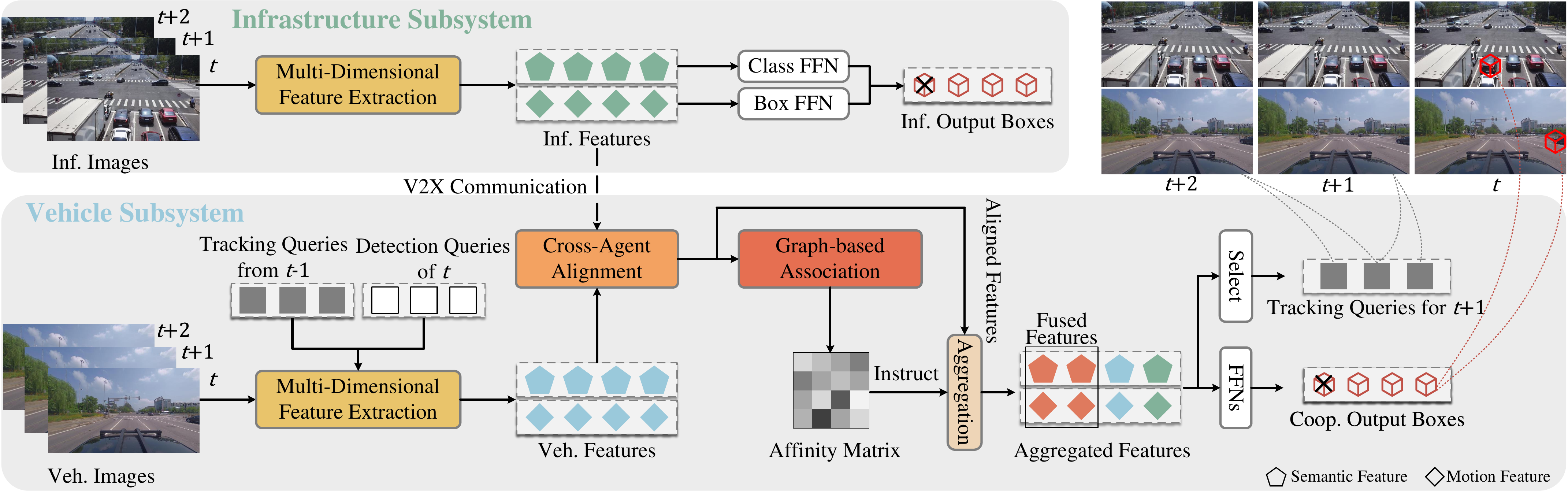}
  \vspace{-7pt}
   \caption{\textbf{Overview of the proposed CoopTrack.}
   In the infrastructure subsystem, sparse instance-level features extracted by the Multi-Dimensional Feature Extraction (MDFE) module are transmitted. Similarly, in the vehicle subsystem, the MDFE module is also employed to extract instance-level features. Upon receiving the infrastructure features, the Cross-Agent Alignment (CAA) module is utilized to transform the infrastructure features to ensure cross-agent consistency. The Graph-Based Association (GBA) module is then used to adaptively associate cross-agent features using a learned affinity matrix, and the Aggregation integrates the two sets of features into one. Finally, the aggregated features are fed into the Select and FFNs to update tracking queries for the next frame and predict the current boxes.
   }
   \label{fig:fig3}
   \vspace{-12pt}
\end{figure*}

\subsection{Cooperative Perception}
Cooperative perception is a hot research topic in autonomous driving, and numerous excellent works have explored this issue \cite{yue2024development,li2024delving}. First, both simulated datasets \cite{mao2022dolphins, li2022v2x, xu2022opv2v, wang2025griffin} and real-world datasets \cite{yu2022dair,xu2023v2v4real,zimmer2024tumtraf,ma2024holovic,hao2024rcooper,yang2024v2x,xiang2024v2x,zhou2024v2xpnp} provide rich data and have made significant contributions to the community. 
Meanwhile, cooperative perception methods have made remarkable progress. Based on the form of the transmitted data, these methods can be divided into early fusion \cite{li2022v2x,xu2022opv2v}, intermediate fusion \cite{xu2022v2x, hu2022where2comm, lin2024v2vformer,yang2023what2comm,hu2024communication,hu2024pragmatic}, and late fusion \cite{yu2022dair}. Among these, intermediate fusion has become a mainstream paradigm due to the flexibility and modeling capabilities of deep features. 

In terms of research scope, the vast majority of previous works focus on single-frame perception tasks, including 3D object detection \cite{xu2022v2x,yang2023what2comm}, BEV segmentation \cite{xu2022cobevt}, and 3D occupancy prediction \cite{song2024collaborative,yan2024pointssc}. In the more challenging field of cooperative sequential perception, several approaches \cite{ruan2025learning,zhou2024v2xpnp,zhang2025co,yin2025knowledge} have attempted to address motion prediction \cite{xu2025component} from a cooperative perspective.
Meanwhile, other studies \cite{xu2023v2v4real, yu2023v2x, chiu2024probabilistic, su2024collaborative, su20233d, su2024cooperative,yu2024end,yang2024lidar} have explored cooperative 3D MOT. Most of them adhere to the tracking-by-cooperative-detection paradigm, which fails to unify tracking with cooperation, resulting in suboptimal performance.

Diverging from existing studies, this paper explores the end-to-end pipeline for cooperative 3D MOT, proposing a novel framework that achieves superior performance.

\subsection{3D Multi-Object Tracking}
3D MOT is a critical perception task for autonomous driving. It involves continuously identifying and locating objects of interest frame by frame from sensor input data \cite{cao2024review}.

In the early stages of research, 3D MOT is often seen as an extension of 3D object detection. Most methods follow the tracking-by-detection paradigm \cite{weng20203d, wang2023camo, yin2021center,pang2022simpletrack,li2022time3d,stearns2022spot,wang2024multi,yang2022quality,weng2020gnn3dmot,sadjadpour2023shasta,marinello2022triplettrack}, which includes modules such as object detection, state estimation, and data association. However, due to the decoupled optimization, such approaches often fail to achieve optimal performance, as the isolated design of modules leads to suboptimal feature learning and error accumulation.
Inspired by MOTR \cite{zeng2022motr}, end-to-end methods \cite{zhou2024ua,cheong2024jdt3d,lin2023sparse4d,li2023end} has become a trend in 3D MOT. This approach leverages object queries to identify and represent instances across frames, achieving superior tracking performance. For example, MUTR3D \cite{zhang2022mutr3d} introduces track queries to locate tracking targets in the current frame, eliminating the need for data association. PF-Track \cite{pang2023standing} proposes a joint tracking and prediction framework that uses historical queries to refine tracking results and leverages prediction outcomes to address occlusion issues. ADA-Track \cite{ding2024ada} introduces decoupled task-dependent queries with a differentiable association module to achieve a better trade-off between detection and tracking tasks. OneTrack \cite{wang2024onetrack} effectively resolves performance conflicts between detection and tracking tasks by coordinating conflicting gradients and dynamically grouping object queries.
Furthermore, some works \cite{luiten2020track} combine 3D MOT with reconstruction \cite{xu2024parameterization,xu2023deformable} to better understand dynamic scenes.

Despite the insights provided by these methods, they have not explored the tracking from a multi-agent perspective.
To bridge this gap, this paper aims to enhance cooperative tracking performance through end-to-end learning.

\section{Method}
\label{sec:method}

This section introduces CoopTrack, whose architecture is illustrated in \cref{fig:fig3}. We begin by presenting the cooperative sequential perception problem in \cref{sec:method-1}, followed by the overall pipeline description. \cref{sec:method-3} and \cref{sec:method-4} detail the key modules, while \cref{sec:method-5} presents the training process.

\subsection{Problem Formulation}
\label{sec:method-1}

Cooperative sequential perception aims to achieve a comprehensive understanding of the surrounding environment over time by integrating the ego vehicle's perception data with information from other agents \cite{yu2023v2x}. During runtime, the system takes as input time-series sensor data from each agent, along with their inter-agent relative poses at each timestamp.
In this paper, we focus on the vehicle-infrastructure cooperative 3D MOT task, which takes images as input and predicts a set of 3D bounding boxes $\mathcal{B}_t$ with consistent IDs across frames, where $t$ is the timestamp. Each box $b_t^i\in\mathcal{B}_t$ is defined by its center, dimension, orientation, velocity as well as a class label, i.e., $b_t^i=[x,y,z,w,l,h,\theta,v_x,v_y]$ \cite{caesarNuScenesMultimodalDataset2020}.

\begin{figure}[t]
  \centering
  \includegraphics[width=0.95\linewidth]{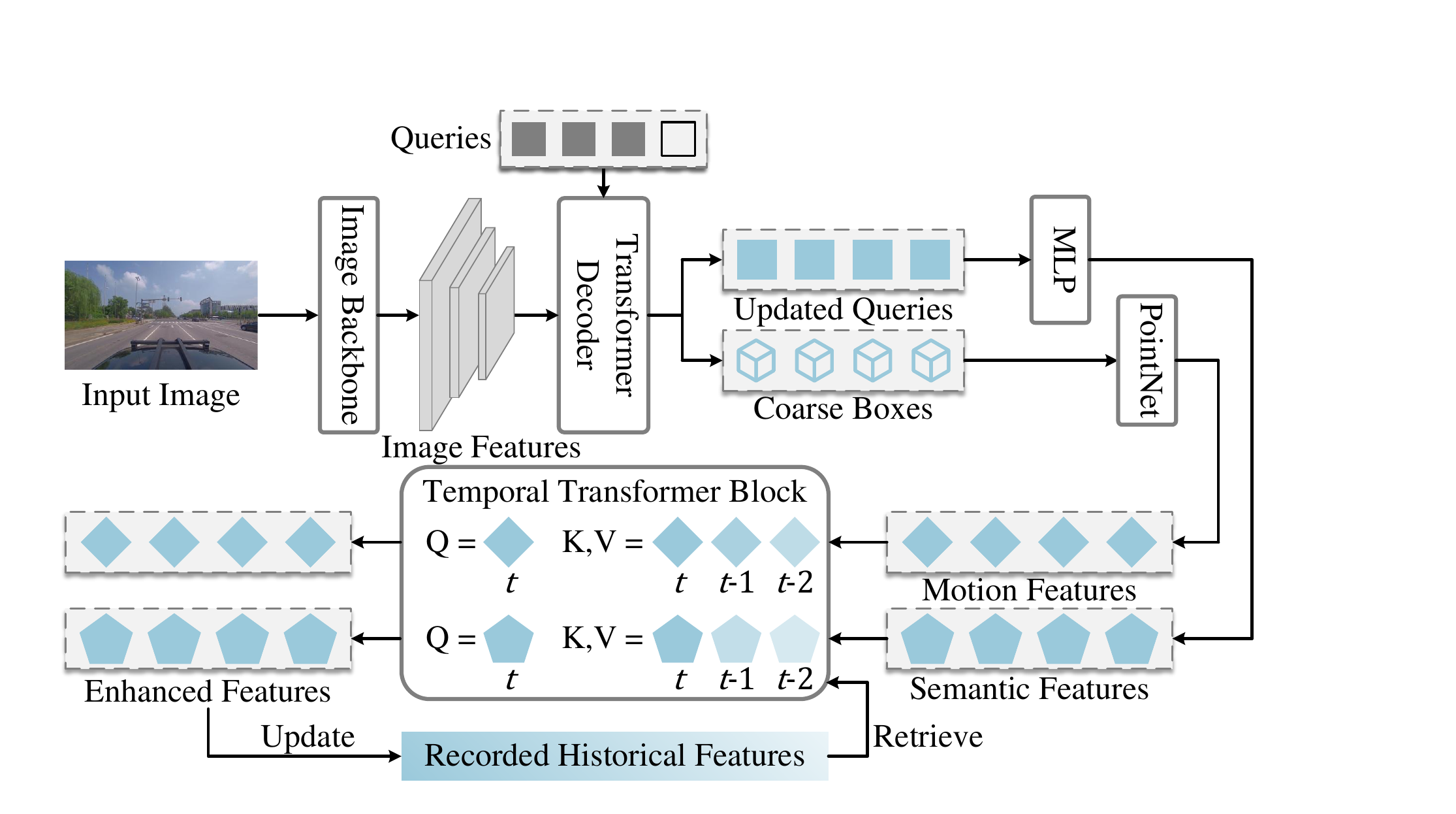}
  \vspace{-7pt}
   \caption{\textbf{Multi-Dimensional Feature Extraction.} The module operates identically for both the vehicle and infrastructure; we use the vehicle side as an example hereafter. First, queries and image features are processed through a transformer decoder, whose output is then used to generate semantic and motion features. Both types of features are then enhanced by interacting with historical information through a temporal transformer block. The recorded historical features are iteratively updated in a sliding window.}
   \vspace{-10pt}
   \label{fig:fig4}
\end{figure}

\subsection{CoopTrack Framework}
\label{sec:method-2}
Our proposed framework excels in the cooperative sequential perception task by maximizing the use of temporal information and cooperation. As illustrated in \cref{fig:fig3}, CoopTrack incorporates three key features: 1) A meticulously designed pipeline, which sets it apart from the existing end-to-end method \cite{yu2024end} as mentioned in \cref{sec:introduction}; 2) Instance-level feature fusion based on a novel feature extraction process, which achieves an optimal trade-off between performance and transmission cost; 3) end-to-end learning equipped with learnable cross-agent alignment and graph-based association, which enhance cooperation.

CoopTrack consists of two subsystems, namely vehicle and infrastructure. Firstly, at each timestamp $t$, the ego vehicle concatenates the tracking queries inherited from the previous timestamp $t-1$ with a fixed number of randomly initialized queries as the queries $\mathcal{Q}_t^{\mathrm{V}}$. Here, the superscript $\mathrm{V}$ represents the vehicle, while the superscript $\mathrm{I}$ in the following text represents the infrastructure.
Each query $q_t^i\in\mathcal{Q}_t$, including a feature vector and a 3D reference point, represents a potential object, i.e., $q_t^i=\{f_t^i\in\mathbb{R}^{1\times d}, p_t^i\in\mathbb{R}^{1\times3}\}$, where $d$ is the feature dimension.
These queries interact with the current image features $F_t^{\mathrm{V}}$ and the historical features $\{\mathcal{M}_{t-\tau:t-1}^{\mathrm{V}},\mathcal{S}_{t-\tau:t-1}^{\mathrm{V}}\}$ in the Multi-Dimensional Feature Extraction (MDFE) module to obtain comprehensive representation of the instances:
\begin{equation}
    \mathcal{M}_t^{\mathrm{V}}, \mathcal{S}_t^{\mathrm{V}} = \mathbf{MDFE}(F_t^{\mathrm{V}}, \mathcal{Q}_t^{\mathrm{V}}, \mathcal{M}_{t-\tau:t-1}^{\mathrm{V}},\mathcal{S}_{t-\tau:t-1}^{\mathrm{V}}),
\end{equation}
where $\mathcal{M}_t^{\mathrm{V}}\in\mathbb{R}^{N_\mathrm{V}\times d}$ and $\mathcal{S}_t^{\mathrm{V}}\in\mathbb{R}^{N_\mathrm{V}\times d}$ are resulting motion features and semantic features, respectively, $N_\mathrm{V}$ represents the number of instances observed by ego and $\tau$ is the length of historical features.

Meanwhile, the infrastructure also undergoes the same process to obtain multi-dimensional features $\mathcal{M}_t^{\mathrm{I}}\in\mathbb{R}^{N_\mathrm{I}\times d}$ and $\mathcal{S}_t^{\mathrm{I}}\in\mathbb{R}^{N_\mathrm{I}\times d}$, where $N_\mathrm{I}$ is the number of instances from infrastructure. 
Note that the MDFE for vehicle and infrastructure are not shared and these features can be decoded into the tracking results of infrastructure by a pair of feed-forward networks (FFN).

Subsequently, these features are transmitted to the vehicle through V2X communication to support cooperation. Then, the Cross-Agent Alignment (CAA) module is used to bridge the domain gap across agents:
\begin{equation}
    \widetilde{\mathcal{M}}_t^{\mathrm{V}}, \widetilde{\mathcal{S}}_t^{\mathrm{V}}, \widetilde{\mathcal{M}}_t^{\mathrm{I}}, \widetilde{\mathcal{S}}_t^{\mathrm{I}} = \mathbf{CAA}(\mathcal{M}_t^{\mathrm{V}}, \mathcal{S}_t^{\mathrm{V}}, \mathcal{M}_t^{\mathrm{I}}, \mathcal{S}_t^{\mathrm{I}},\mathbf{R},\mathbf{t}),
\end{equation}
where $\widetilde{\mathcal{M}}_t^{\mathrm{V}}, \widetilde{\mathcal{S}}_t^{\mathrm{V}}, \widetilde{\mathcal{M}}_t^{\mathrm{I}}, \widetilde{\mathcal{S}}_t^{\mathrm{I}}$ denote the aligned features and $\mathbf{R}\in\mathbb{R}^{3\times3},\mathbf{t}\in\mathbb{R}^{1\times3}$ are rotation and translation, respectively. Next, the Graph-based Association (GBA) module learns the relationship between ego and roadside instances from the aligned features, resulting in an affinity matrix $A_t\in\mathbb{R}^{N_{\mathrm{V}}\times {N_{\mathrm{I}}}}$:
\begin{equation}
    A_t = \mathbf{GBA}(\widetilde{\mathcal{M}}_t^{\mathrm{V}}, \widetilde{\mathcal{S}}_t^{\mathrm{V}}, \widetilde{\mathcal{M}}_t^{\mathrm{I}}, \widetilde{\mathcal{S}}_t^{\mathrm{I}}).
\end{equation}
Each element $a_{i,j}\in A_t$ represents the similarity between the $i$-th vehicle-side instance and the $j$-th roadside instance. The closer the value is to 1, the more likely they correspond to the same instance.
Instructed by the affinity matrix, multiagent features are aggregated into a new set of instance features $\{\mathcal{M}_t^{\mathrm{R}}, \mathcal{S}_t^{\mathrm{R}}\}$:
\begin{equation}
    \mathcal{M}_t, \mathcal{S}_t = \mathbf{Aggr}(A_t, \widetilde{\mathcal{M}}_t^{\mathrm{V}}, \widetilde{\mathcal{S}}_t^{\mathrm{V}}, \widetilde{\mathcal{M}}_t^{\mathrm{I}}, \widetilde{\mathcal{S}}_t^{\mathrm{I}}).
\end{equation}

Finally, two FFNs are used to decode the location, dimension and motion states of objects from the motion features and the category based on the semantic features, respectively. These two parts form the output of CoopTrack.

To achieve continuous tracking across frames, we select active instances and propagate them to the next frame $t+1$. For propagated queries, semantic features are directly utilized as the feature vectors, while the constant velocity assumption \cite{zhang2022mutr3d} is adopted to predict reference points.

\subsection{Multi-Dimensional Feature Extraction}
\label{sec:method-3}
CoopTrack employs instance-level features to represent potential objects in the scene, enabling information sharing both across agents and modules within each frame and temporally between frames.
This end-to-end cooperative tracking framework necessitates comprehensive instance modeling.
While query-based methods \cite{zhang2022mutr3d,ding2024ada,fan2024quest} demonstrate effectiveness, we identify two key limitations: 1) their dependence on single-frame features compromises temporal modeling, and 2) the implicit coupling of semantic and motion information introduces decoding ambiguity.
To address these issues, we decouple 2D semantic features from 3D motion features, constructing multi-dimensional instance representations that are subsequently refined through temporal interaction.

This process is consistent for both vehicles and roads, therefore the superscripts are omitted. 
As shown in \cref{fig:fig4}, at the timestamp $t$, queries $\mathcal{Q}_t$ first interact with image features $F_t$ extracted by the image backbone in the transformer decoder, resulting in the coarse 3D bounding boxes $\hat{\mathcal{B}}_t$ and updated queries $\hat{\mathcal{Q}}_t$.
Note that this process is agnostic to the query-based detector, and we adopt a classical detection method, BEVFormer \cite{li2024bevformer}, for experiments.

Subsequently, the multi-dimensional features are extracted. A multi-layer perceptron (MLP) is employed to extract semantic features $\mathcal{S}_t$ from query features $\hat{\mathcal{Q}}_t$.
Meanwhile, 3D corners $\mathcal{C}_t^i\in\mathbb{R}^{N\times8\times3}$ are obtained from coarse 3D bounding boxes $\hat{\mathcal{B}}_t$. Since we are more interested in the geometry of each instance, we calculate the relative coordinates of each corner to their center. These corners are then flattened and fed into PointNet \cite{qi2017pointnet}, which consists of a 4-layer MLP and a max-pooling layer, to obtain the motion features $\mathcal{M}_t\in\mathbb{R}^{N\times d}$.

Inspired by PF-Track \cite{pang2023standing}, a specialized temporal transformer block, which contains 2 decoder layers, is introduced to empower these features with temporal awareness.
Firstly, we enhance features through sinusoidal positional encoding \cite{vaswani2017attention} to capture temporal dependencies. To handle variable-length sequences, shorter instance histories are zero-padded and accompanied by binary masking, effectively preventing interaction with padded positions during attention computation.
At last, the historical features are updated in a first-in-first-out manner.

With these multi-dimensional representations encompassing both semantic and motion features, the distinctiveness of different instances and the consistency of the same instance can be ensured, thereby reducing the difficulty of subsequent association and fusion.

\begin{figure}[t]
  \centering
  \includegraphics[width=0.9\linewidth]{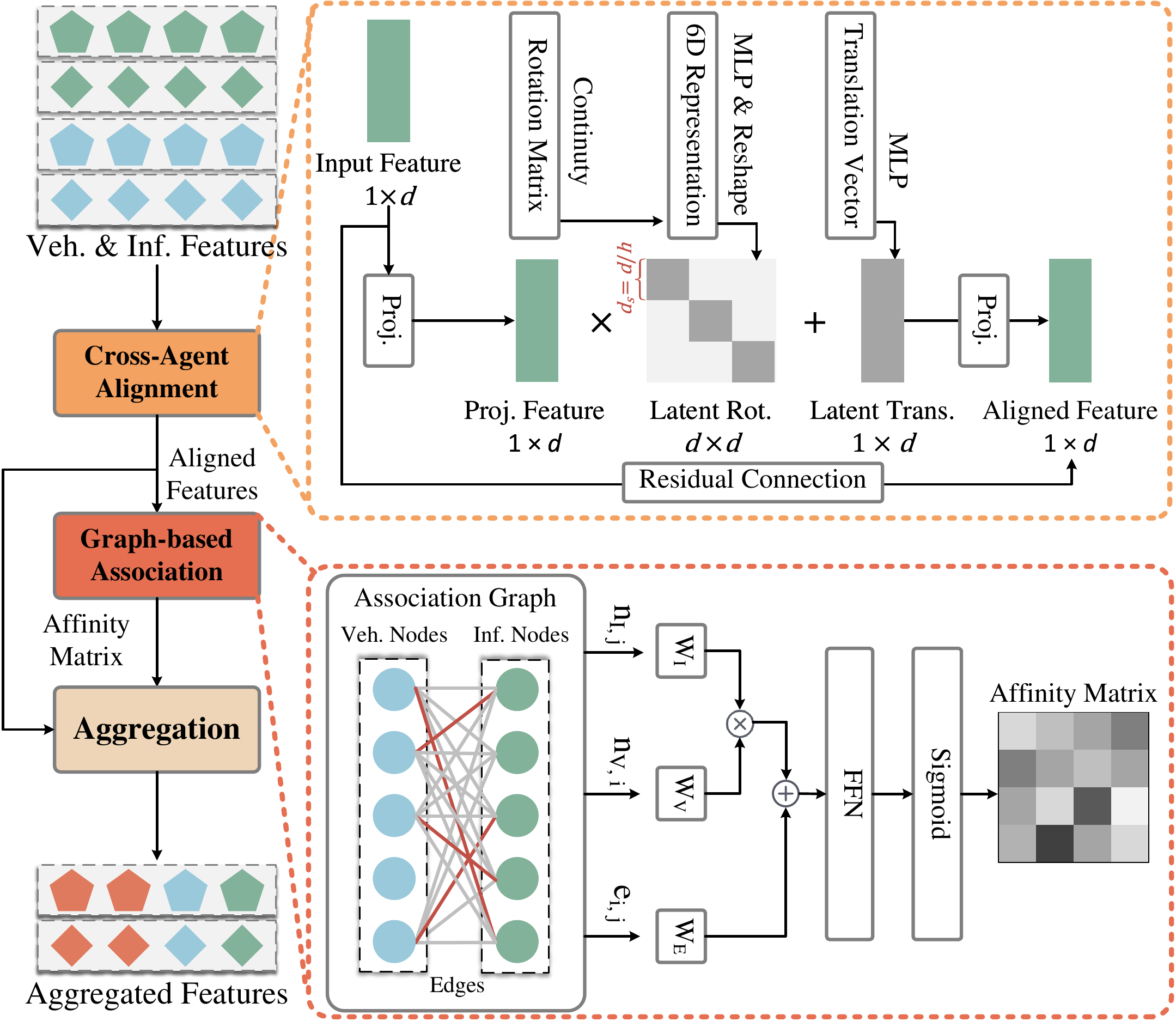}
  \vspace{-7pt}
   \caption{\textbf{Cross-Agent Association and Aggregation.} The Cross-Agent Alignment module first addresses the vehicle-infrastructure domain gap via latent space transformation. Using graph attention, the Graph-based Association module then learns cross-agent instance relationships. The Aggregation module then adaptively fuses the aligned and associated features.}
   \vspace{-10pt}
   \label{fig:fig5}
\end{figure}

\subsection{Cross-Agent Association and Aggregation}
\label{sec:method-4}
This subsection explains how to fuse instance-level features from different agents. Intuitively, we aim to merge features that describe the same instance into one to refine the representation; for features that represent distinctive instances, we use them directly to broaden the observation. Therefore, cross-agent association is first performed. Then, based on the matching results, features from different agents are aggregated for decoding cooperative outputs.

However, attempting to directly establish an association between the received features and those of the ego agent would be less than ideal. This suboptimal approach is due to the inherent differences in sensors, perspectives, and spatial positions between the two agents, which inevitably give rise to a domain gap in their respective feature sets. Therefore, the cross-agent alignment module is introduced.

\begin{table*}[tbp]
    \centering
    \renewcommand\arraystretch{0.9}
    \scalebox{0.95}
    {
    \begin{tabular}{c|c|cccc}
    \toprule
    \textbf{Method} & \textbf{Paradigm} & \textbf{mAP} $ \uparrow$ & \textbf{AMOTA} $ \uparrow$ & \textbf{Trans. Cost} $\downarrow$ \\
    \midrule
    No Fusion \cite{li2024bevformer} & - & 0.165 & 0.163 & 0 \\
    BEV Feature Fusion \cite{yu2023v2x} & TBCD & 0.243 & 0.209 & 8.19$\times 10^7$ \\
    CoAlign \cite{lu2023robust} & TBCD & 0.240 & 0.234 & 8.19$\times 10^7$ \\
    CoCa3D \cite{hu2023collaboration}  & TBCD & 0.226 & - & 4.63$\times 10^6$ \\
    V2X-ViT \cite{xu2022v2x} & TBCD & 0.268 & \underline{0.287} & 2.56$\times 10^6$ \\
    Where2comm \cite{hu2022where2comm} & TBCD & 0.162 & 0.106 & 5.40$\times 10^5$ \\
    DiscoNet \cite{li2021learning} & TBCD & 0.216 & 0.203 & 1.60$\times 10^5$ \\
    \hline
    V2X-ViT \cite{xu2022v2x}+Where2comm \cite{hu2022where2comm}  & TBCD & 0.178 & 0.071 & 7.22$\times 10^4$ \\
    Late Fusion \cite{yu2023v2x} & TBCD & 0.196 & 0.263  & \textbf{6.60}$\mathbf{\times 10^2}$ \\
    UniV2X \cite{yu2024end} & E2EC & \underline{0.295} & 0.239 & 6.96$\times 10^4$ \\
    \rowcolor[gray]{.9} 
    CoopTrack (Ours) & E2EC &\textbf{0.390}~\textcolor{red}{(+0.095)} & \textbf{0.328}~\textcolor{red}{(+0.041)}& $\underline{5.64\times 10^4}$& \\
    \bottomrule
    \end{tabular}
    }
    \vspace{-7pt}
    \caption{\textbf{Perception Performance and Transmission Cost Comparison on the V2X-Seq dataset.} TBCD denotes the tracking-by-cooperative-detection paradigm, and E2EC denotes end-to-end cooperative tracking. \textbf{Bold} indicates the best results, and \underline{underlining} indicates the second-best results. \textcolor{red}{Red} denotes the improvement compared to the second-best results.}
    \vspace{-10pt}
    \label{tab:comparison_2hz}
\end{table*}

\noindent\textbf{Cross-Agent Alignment.} As revealed in \cite{ruppel2022transformers,doll2023star}, the impact of the aforementioned differences on the latent space can be regarded as a linear operator. It can be represented as a process of latent transformation applied to the source feature vectors (\cref{equ:equ1}), akin to the explicit spatial transformation applied to reference points (\cref{equ:equ2}):
\begin{subequations}
    \begin{align}
        \widetilde{\mathcal{M}}^{\mathrm{I}} = \mathcal{M}^{\mathrm{I}} \cdot \hat{\mathbf{R}}^\top + \hat{\mathbf{t}}, \label{equ:equ1} \\
        \widetilde{\mathcal{P}}^{\mathrm{I}} = \mathcal{P}^{\mathrm{I}} \cdot \mathbf{R}^\top + \mathbf{t}, \label{equ:equ2}
    \end{align}
\end{subequations}
where $\hat{\mathbf{R}}\in\mathbb{R}^{d\times d}$ and $\hat{\mathbf{t}}\in\mathbb{R}^{1\times d}$ denote the learned latent rotation matrix and latent translation vector, while $\mathbf{R}\in\mathbb{R}^{3\times3}$ and $\mathbf{t}\in\mathbb{R}^{1\times3}$ are the explicit counterparts. Note that we take the motion features $\mathcal{M}^{\mathrm{I}}$ as an example, while the semantic features $\mathcal{S}^{\mathrm{I}}$ follow the same pipeline. For simplicity, we omit the subscript of timestamp here.

Next, we delve into the process of acquiring the latent rotation matrix $\hat{\mathbf{R}}$ and the latent translation vector $\hat{\mathbf{t}}$. Specifically, as shown in the upper part of \cref{fig:fig5}, we convert the rotation matrix $\mathbf{R}\in\mathbb{R}^{3\times3}$ into a 6D rotation vector through continuous representation \cite{zhou2019continuity}, while the translation vector is preserved. Subsequently, two MLPs are utilized to predict the parameters of the latent rotation and the latent translation, respectively.
We adopt the segmented mapping approach \cite{doll2023star} to reduce the number of learnable parameters and lower training difficulty. Refer to \cite{doll2023star} for more details.

After this process, aligned features are obtained, ensuring cross-agent consistency for the same instance, and thus preparing for subsequent association and fusion.

\begin{table}[t]
\centering
\renewcommand\arraystretch{0.9}
\setlength{\tabcolsep}{4pt} 
\begin{tabular}{cccc}
\toprule
\textbf{Method} & \textbf{mAP}$\uparrow$ & \textbf{AMOTA}$\uparrow$ & \textbf{Trans. Cost}$\downarrow$ \\
\midrule
    No Fusion     &  0.375    &   0.365   &   0\\
    Late Fusion       &    0.378   &    0.377  & $\mathbf{1.56\times10^3}$  \\
Early Fusion      &  \textbf{0.607}    &   \textbf{0.670}   &   $3.11\times10^8$ \\
\midrule
UniV2X \cite{yu2024end}      &     0.419  & 0.456   &   $\underline{5.58\times10^4}$ \\
\rowcolor[gray]{.9} 
CoopTrack (Ours) & \underline{0.479} & \underline{0.488} & $1.17\times10^5$ \\
\bottomrule
\end{tabular}
\vspace{-5pt}
\caption{\textbf{Perception Performance and Transmission Cost Comparison on the Griffin dataset.} \textbf{Bold} indicates the best results, and \underline{underlining} indicates the second-best results.}
\vspace{-11pt}
\label{tab:comparison_griffin}
\end{table}
\begin{table*}[t]
\centering
\renewcommand\arraystretch{0.9}
\setlength{\tabcolsep}{3.0pt}
\scalebox{0.95}

\begin{tabular}{c|c|ccccc|ccc|c}
\toprule
\textbf{Method}          & \textbf{Paradigm} & \textbf{mAP}$\uparrow$   & \textbf{ATE}$\downarrow$   & \textbf{ASE}$\downarrow$   & \textbf{AOE}$\downarrow$   & \textbf{AVE}$\downarrow$   & \textbf{AMOTA}$\uparrow$ & \textbf{AMOTP}$\downarrow$ & \textbf{IDS}$\downarrow$ & \textbf{Trans. Cost}$\downarrow$ \\
\midrule
\multicolumn{11}{c}{ResNet50}                                                                           \\
\midrule
No Fusion       & -        & 0.110 & 0.946 & 0.153 & 0.198 & 2.989 & 0.087 & 1.796  & \textbf{27}  & 0   \\
UniV2X-Track    & E2EC     & 0.310 & 0.922 & \textbf{0.147} & 0.232 & 2.702 & 0.266 & 1.480   & 711 & \textbf{1.57}$\mathbf{\times10^5}$  \\
\rowcolor[gray]{.9} 
CoopTrack(Ours) & E2EC     & \textbf{0.356} & \textbf{0.777} & 0.154 & \textbf{0.117} & \textbf{2.594} & \textbf{0.346} & \textbf{1.376}  & 201 & 2.85$\times10^5$  \\
\midrule
\midrule
\multicolumn{11}{c}{ResNet101}                                                                          \\
\midrule
No Fusion       & -        & 0.183 & 0.810 & 0.145 & 0.153 & 2.687 & 0.217 & 1.658  & \textbf{50}  & 0   \\
LF+AB3DMOT      & TBCD      &  0.419   &  \textbf{0.627}  &  0.148   &  0.125  &   \textbf{2.048}   & 0.371   &  1.395   & 620 & \textbf{3.52}$\mathbf{\times10^3}$   \\
BFF+AB3DMOT     & TBCD      &  0.430   &  0.631   &  0.158  &  0.221   &  2.282   &  0.390   &  1.375   & 565   & 4.10$\times10^8$   \\
UniV2X-Track    & E2EC     & 0.394 & 0.739 & 0.138 & 0.162 & 2.311 & 0.418 & \textbf{1.233}  & 535 & 1.51$\times10^5$   \\
\rowcolor[gray]{.9} 
CoopTrack (Ours) & E2EC     & \textbf{0.442} & 0.715 & \textbf{0.135} & \textbf{0.108} & 2.466 & \textbf{0.435} & 1.253  & 302 & 2.88$\times10^5$ \\
\bottomrule
\end{tabular}
\vspace{-5pt}
\caption{\textbf{Perception Performance and Transmission Cost Comparison on the 10Hz V2X-Seq dataset.} LF denotes Late Fusion, and BFF indicates BEV Feature Fusion. \textbf{Bold} indicates the best results.}
\vspace{-7pt}
\label{tab:comparison}
\end{table*}

\noindent\textbf{Graph-based Association.}
In instance-level cooperative perception, the accurate association of instance features is of critical importance. Firstly, failure to correctly match features that describe the same instance can lead to duplicate detections, thereby increasing the number of false positives. Secondly, the erroneous association of features belonging to two distinct instances may result in missed detections and even trigger identity switches. 
Previous cooperative perception methods primarily rely on explicit spatial relative positions for association, such as Euclidean distances of reference points  \cite{fan2024quest,zhong2024leveraging,yu2024end} and relative position embeddings \cite{chen2023transiff}. They suffer from insufficient information and noisy relative pose, which prevent accurate association and impair the final cooperative performance. Therefore, inspired by \cite{li2022time3d,ding2024ada}, we attempt to learn the relationships between instances from multi-dimensional features, which contain more information compared to reference points and position embeddings.

As shown in the lower part of \cref{fig:fig5}, we adopt the graph-based attention \cite{ding20233dmotformer} to learn the affinity matrix $A$.
Firstly, a fully connected association graph $\mathcal{G}=\{\mathcal{N}, \mathcal{E}\}$ is constructed between agents by combining the multi-dimensional features and 3D reference points. For the ego vehicle, the node features $\mathcal{N}^{\mathrm{V}}\in\mathbb{R}^{N_\mathrm{V}\times d}$ are generated by applying an MLP on the concatenation of motion features and semantic features. Meanwhile, the node features $\mathcal{N}^{\mathrm{I}}\in\mathbb{R}^{N_\mathrm{I}\times d}$ of infrastructure are obtained in the same manner. 
Another MLP produces the edge features $\mathcal{E}\in\mathbb{R}^{N_{\mathrm{V}}\times N_{\mathrm{I}}\times d}$ with the input of the pair-wise differences $\mathcal{D}\in\mathbb{R}^{N_{\mathrm{V}}\times N_{\mathrm{I}}\times 3}$. Each element $d_{i,j}\in\mathcal{D}$ represents the distance between the reference points of the $i$-th vehicle node and the $j$-th infrastructure node: $d_{i,j}=\vert p^{\mathrm{V}}_i - p^{\mathrm{I}}_j \vert$.
In the graph-based attention, the dot product of node features is computed, then the edge features are summed:
\begin{equation}
    \hat{A} = \frac{\left( \mathcal{N}^{\mathrm{V}}W^{\mathrm{V}} \right) \left( \mathcal{N}^{\mathrm{I}}W^{\mathrm{I}} \right) ^{\mathrm{T}}}{\sqrt{d}} + \mathcal{E}W^{\mathrm{E}},
\end{equation}
where $W^{\mathrm{V}}, W^{\mathrm{I}}, W^{\mathrm{E}}\in\mathbb{R}^{d\times d}$ are learnable projection weights, and $\hat{A}$ represents the computed attention matrix.
Subsequently, a FNN followed by a sigmoid function predicts the final affinity matrix $A$.

In summary, this learning-based method considers both feature and spatial information, estimates the similarity relationships between instances, and provides guidance for instance-level feature aggregation. Following previous works \cite{fan2024quest, yu2024end}, we also use the Hungarian algorithm \cite{kuhn1955hungarian} to obtain pairs for fusion, with the cost matrix being \(1-A\). For details, please refer to UniV2X \cite{yu2024end}.

\subsection{Training}
\label{sec:method-5}
CoopTrack is trained via a two-stage strategy. 
In the first stage, we train the vehicle-side and infrastructure-side end-to-end tracking models separately using their respective ground truth. We use Focal Loss for classification with $\alpha=0.25$ and $\gamma=2.0$, and L1 Loss for regression loss. The overall loss can be formulated as follows:
\begin{equation}
    \mathcal{L}_{\mathrm{stage1}} = \lambda_{\mathrm{bbx}}\mathcal{L}_{\mathrm{bbx}}+\lambda_{\mathrm{cls}}\mathcal{L}_{\mathrm{cls}},
\end{equation}
where $\lambda_{\mathrm{bbx}}=0.25$ and $\lambda_{\mathrm{cls}}=2.0$ are respective loss weights. 
For the second stage, we train end-to-end cooperative tracking and association under the supervision of cooperative ground truth and generated association label, with the pretrained model in the first stage. We treat the association as a binary classification problem and employ Focal Loss for supervision, with $\alpha=0.5$ and $\gamma=1.0$. The overall loss can be formulated as follows:
\begin{equation}
    \mathcal{L}_{\mathrm{stage2}} = \lambda_{\mathrm{bbx}}\mathcal{L}_{\mathrm{bbx}}+\lambda_{\mathrm{cls}}\mathcal{L}_{\mathrm{cls}}+\lambda_{\mathrm{asso}}\mathcal{L}_{\mathrm{asso}},
\end{equation}
where $\lambda_{\mathrm{bbx}}=0.25$, $\lambda_{\mathrm{cls}}=2.0$, $\lambda_{\mathrm{asso}}=10.0$.

\begin{figure}[t]
  \centering
  \includegraphics[width=0.91\linewidth]{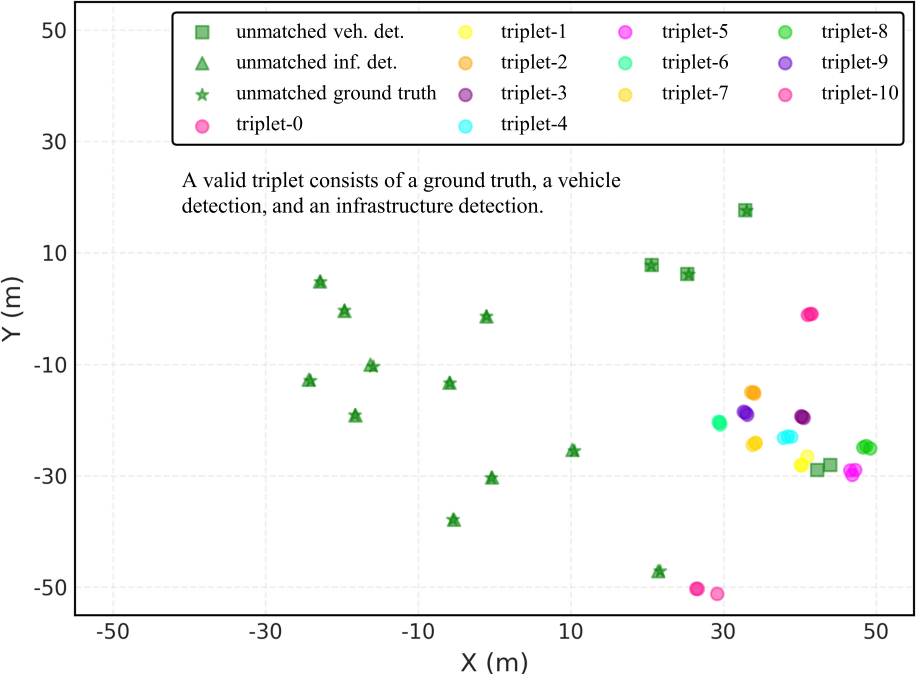}
  \vspace{-7pt}
   \caption{\textbf{The Association Label Generation Process.} We visualize this process by projecting predicted instances and ground truth into the bird's eye view space.}
   \vspace{-10pt}
   \label{fig:fig6}
\end{figure}

In the second stage, to generate association labels, we use the Hungarian algorithm \cite{kuhn1955hungarian} to match cross-agent instances with ground truth based on the L1 cost of their predicted attributes. If a ground truth successfully matches with instances from both the vehicle and infrastructure, the corresponding position in the label matrix is set to positive. For other cases, including instances that do not match any ground truth or instances that match ground truth with different IDs, the corresponding position is set to negative. We visualize the label generation process in \cref{fig:fig6}. The results demonstrate that the first-stage model achieves competent tracking performance, thereby validating the reliability of the generated labels.

\section{Experiments}
\label{sec:experiments}

We evaluate our approach on two benchmark datasets: V2X-Seq \cite{yu2023v2x} and Griffin \cite{wang2025griffin}. Additional experimental details, more ablation studies, and qualitative analyses are provided in the appendix.

\subsection{Experiment Setup}
\label{sec:experiment setup}
\textbf{Datasets.} \textbf{V2X-Seq} \cite{yu2023v2x} is a real-world dataset specifically designed for vehicle-infrastructure cooperative perception. The dataset comprises approximately 100 dynamic scenes, each lasting 10-25 seconds at 10Hz frequency. We adopt the official data split and evaluate on both the native 10Hz data and a 2Hz downsampled variant, following UniV2X \cite{yu2024end}, for fair comparison with existing methods.
\textbf{Griffin} \cite{wang2025griffin} focuses on aerial-ground cooperative perception and includes multi-view images of cooperative agents collected in Carla. We evaluate CoopTrack on the Griffin-25m dataset, adhering to the official training and validation splits.

\noindent\textbf{Implementation.} For V2X-Seq, we set the horizontal perception range of the ego-vehicle to [-51.2, 51.2, -51.2, 51.2] meters, and that of the roadside unit to [0, 102.4, -51.2, 51.2] meters, with both having a height range of [-5, 3] meters. For Griffin, we keep the horizontal perception range consistent for vehicles and UAVs, set to [-51.2, 51.2, -51.2, 51.2] meters, while the height perception range of UAVs is [-35, -20] meters.
We adopt BEVFormer \cite{li2024bevformer} as the detector with two backbone sizes: ResNet50 and ResNet101 \cite{he2016deep}. We conduct all ablation studies using the ResNet50.
We initialize the corresponding models using a pre-trained single-frame detection model trained for 24 epochs. We use the AdamW optimizer with a weight decay of 0.01 to train the models of the proposed framework. The initial learning rate is set to $2\times10^{-4}$, and cosine annealing is applied to the learning rate schedule. All experiments are conducted on NVIDIA 3090 GPUs. 

\subsection{Comparison with Existing Works}
\textbf{V2X-Seq Results.} As depicted in \cref{tab:comparison_2hz}, we benchmark the perception performance against other state-of-the-art (SOTA) methods using a ResNet101 backbone, adhering to the configuration of UniV2X. \cite{yu2024end}. Initially, CoopTrack shows enhancements over the No Fusion approach, signifying that cooperation can substantially elevate performance by information fusion.
In comparison to other methods, our approach achieves superior performance at a lower transmission cost. For instance, relative to the excellent work, V2X-ViT \cite{xu2022v2x}, our method boosts mAP by 12.2\% and AMOTA by 4.1\%, while the transmission volume is only 2.2\% of its amount. This indicates that we have reached a new SOTA performance. Although Late Fusion incurs the lowest transmission cost, our method significantly outperforms it, with mAP and AMOTA exceeding 19.4\% and 6.5\%, respectively. These findings demonstrate that our approach achieves the best trade-off between performance and transmission cost through instance-level feature fusion.

Regarding the paradigm, CoopTrack not only surpasses all tracking-by-cooperative-detection methods but also, compared to its end-to-end counterpart (UniV2X), achieves a 9.5\% increase in mAP and 8.9\% in AMOTA under similar transmission costs. This suggests that 1) end-to-end tracking, benefiting from unified training, can better transmit upstream detection information to the downstream tracking task, thereby enhancing performance; 2) the pipeline we propose, along with learning-based association, can more effectively integrate the cooperative module with end-to-end tracking, fully realizing the benefits of cooperative information for the tracking task.

\noindent\textbf{Griffin Results.} As shown in \cref{tab:comparison_griffin}, CoopTrack demonstrates consistent performance advantages in aerial-ground cooperation scenarios. The framework achieves better perception accuracy than the No Fusion baseline, confirming its adaptability to different sensor configurations and scenarios. Compared to both Late Fusion and Early Fusion, CoopTrack establishes superior performance-bandwidth trade-offs. Most notably, it outperforms UniV2X \cite{yu2024end} by 6.0\% in mAP and 3.2\% in AMOTA with similar transmission costs, validating the effectiveness of our approach.

\begin{table}[t]
\centering
\renewcommand\arraystretch{0.9}
\setlength{\tabcolsep}{4pt} 
\begin{tabular}{cccc|cc}
\toprule
\textbf{Pipeline} & \textbf{MDFE} & \textbf{CAA} & \textbf{GBA} & \textbf{mAP}$\uparrow$   & \textbf{AMOTA}$\uparrow$ \\
\midrule
         &      &     &     & 0.310 & 0.266 \\
\checkmark       &      &     &     & 0.337 & 0.277 \\
\checkmark        & \checkmark    &     &     & 0.345 & 0.283 \\
\checkmark       &      & \checkmark   &     & 0.354 & 0.304 \\
\checkmark       &      &     & \checkmark   &   0.342    & 0.295    \\
\checkmark       & \checkmark  & \checkmark  &     & 0.355 & 0.332 \\
\rowcolor[gray]{.9} 
\checkmark      & \checkmark   & \checkmark  & \checkmark & \textbf{0.356} & \textbf{0.346} \\
\bottomrule
\end{tabular}
\vspace{-5pt}
\caption{\textbf{Ablation Study of Each Module.} MDFE is the multi-dimensional feature extraction, CAA represents the cross-agent alignment and GBA is the graph-based association.}
\vspace{-9pt}
\label{tab:ablation_module}
\end{table}

\subsection{Ablation Study}

\noindent\textbf{More Results on V2X-Seq.} As presented in \cref{tab:comparison}, we reproduce two classical methods and UniV2X-Track \cite{yu2024end} on the 10Hz dataset.
Firstly, compared to \cref{tab:comparison_2hz}, a higher frequency helps achieve better tracking performance. For example, with the same ResNet101 backbone, CoopTrack shows an improvement of 10.7\% in AMOTA. This is reasonable as, on one hand, the amount of training data increases, and on the other hand, the shorter frame interval facilitates predicting the next frame's position.

Consistent with \cref{tab:comparison_2hz}, we find that end-to-end cooperative tracking significantly outperforms the tracking-by-cooperative-detection approaches. 
Compared to UniV2X, CoopTrack shows a clear advantage. With the ResNet50 backbone, it achieves 35.6\% mAP and 34.6\% AMOTA, improving by 4.6\% mAP and 8.0\% AMOTA compared to UniV2X-Track. This indicates that the proposed pipeline and learning-based association are more suitable for the cooperative tracking. Meanwhile, the performance of the two backbones also demonstrates the scalability of our method.

\noindent\textbf{Effect of Each Module.}
As shown in \cref{tab:ablation_module}, we incrementally add modules based on the baseline to demonstrate the effect of each proposed module. For the proposed pipeline, we achieve an improvement of 2.7\% mAP and 1.1\% AMOTA, indicating that cutting off the interaction between other agents’ queries and the ego vehicle’s features to avoid ambiguity and conflict can enhance detection and tracking performance.
Individually adding the three key modules, MDFE, CAA, and GBA, all result in improved detection and tracking performance, demonstrating their crucial role in cooperative tracking. Additionally, the combination of these modules further boosts performance. The joint effect of MDFE and CAA achieves 35.5\% mAP and 33.2\% AMOTA, and adding the GBA module on this basis further gains an improvement of 1.4\% AMOTA, suggesting that learning-based association can effectively benefit cooperative tracking.

\begin{table}[t]
\centering
\renewcommand\arraystretch{0.9}
\setlength{\tabcolsep}{4pt} 
\begin{tabular}{c|ccccc}
\toprule
Number & mAP$\uparrow$   & AMOTA$\uparrow$ & AMOTP$\downarrow$ & MT$\uparrow$ & ML$\downarrow$ \\
\midrule
0      & 0.132 & 0.100 & 1.739 & 41  & 519 \\
1      & 0.287 & 0.123 & 1.402 & 78  & 417 \\
2      & 0.300 & 0.174 & 1.398 & 80  & 425 \\
3      & 0.332 & 0.255 & 1.414 & 103 & 361 \\
\rowcolor[gray]{.9}
4      & 0.356 & \textbf{0.346} & \textbf{1.376} & \textbf{143} & 350\\
5      & \textbf{0.367} & 0.324 & 1.409 & 136 & \textbf{339}\\   

\bottomrule
\end{tabular}
\vspace{-5pt}
\caption{\textbf{Influence of the Number of Historical Frames.}}
\vspace{-2pt}
\label{tab:history}
\end{table}

\noindent\textbf{Number of Historical Features.}
We investigated the influence of the number of historical features in the multi-dimensional feature extraction module. As the number of historical frames increases from 0 to 4, we observe an improvement in both detection and tracking performance in \cref{tab:history}, indicating that temporal cues are beneficial for both tasks. The performance begins to saturate at 4 frames. This is due to our iterative approach of propagating track queries between frames and updating the feature queue, which effectively endows these features with a long historical receptive field, thereby eliminating the need for interaction with distant historical features. Considering the balance between efficiency and performance, we opt to use 4 frames to enhance the current features.
\section{Conclusion}
\label{sec:conclusion}
This paper tackles cooperative 3D multi-object tracking by proposing CoopTrack, a novel instance-level fully end-to-end framework featuring learnable instance association and a fusion-after-decoding pipeline.
Experiments on the V2X-Seq and Griffin datasets show that CoopTrack achieves excellent performance and can adapt to various cooperative scenarios and different sensor configurations.
These results validate CoopTrack's effectiveness, yet its limitations merit note: the training process is relatively complex, and future work should explore more efficient training approaches.

\newpage
\section*{Acknowledgments}
We sincerely thank the Wuxi Research Institute of Applied Technologies at Tsinghua University for supporting this work under Grant No. 20242001120. This work is also supported by funding from Horizon Robotics.
{
    \small
    \bibliographystyle{ieeenat_fullname}
    \bibliography{main}

\begin{thebibliography}{81}
\providecommand{\natexlab}[1]{#1}
\providecommand{\url}[1]{\texttt{#1}}
\expandafter\ifx\csname urlstyle\endcsname\relax
  \providecommand{\doi}[1]{doi: #1}\else
  \providecommand{\doi}{doi: \begingroup \urlstyle{rm}\Url}\fi

\bibitem[Caesar et~al.(2020)Caesar, Bankiti, Lang, Vora, Liong, Xu, Krishnan, Pan, Baldan, and Beijbom]{caesarNuScenesMultimodalDataset2020}
Holger Caesar, Varun Bankiti, Alex~H. Lang, Sourabh Vora, Venice~Erin Liong, Qiang Xu, Anush Krishnan, Yu Pan, Giancarlo Baldan, and Oscar Beijbom.
\newblock nuscenes: A multimodal dataset for autonomous driving.
\newblock In \emph{2020 IEEE/CVF Conference on Computer Vision and Pattern Recognition (CVPR)}, pages 11618--11628, 2020.

\bibitem[Cao et~al.(2024)Cao, Zhang, Jin, Lv, Hou, and Zhang]{cao2024review}
Jingwei Cao, Hongyu Zhang, Lisheng Jin, Jiawang Lv, Guoyang Hou, and Chengtao Zhang.
\newblock A review of object tracking methods: From general field to autonomous vehicles.
\newblock \emph{Neurocomputing}, page 127635, 2024.

\bibitem[Chen et~al.(2023)Chen, Shi, and Jia]{chen2023transiff}
Ziming Chen, Yifeng Shi, and Jinrang Jia.
\newblock Transiff: An instance-level feature fusion framework for vehicle-infrastructure cooperative 3d detection with transformers.
\newblock In \emph{Proceedings of the IEEE/CVF International Conference on Computer Vision}, pages 18205--18214, 2023.

\bibitem[Cheong et~al.(2024)Cheong, Zhou, and Waslander]{cheong2024jdt3d}
Brian Cheong, Jiachen Zhou, and Steven Waslander.
\newblock Jdt3d: Addressing the gaps in lidar-based tracking-by-attention.
\newblock In \emph{European Conference on Computer Vision}, pages 161--177. Springer, 2024.

\bibitem[Chiu et~al.(2024)Chiu, Wang, Chen, and Smith]{chiu2024probabilistic}
Hsu-Kuang Chiu, Chien-Yi Wang, Min-Hung Chen, and Stephen~F Smith.
\newblock Probabilistic 3d multi-object cooperative tracking for autonomous driving via differentiable multi-sensor kalman filter.
\newblock In \emph{2024 IEEE International Conference on Robotics and Automation (ICRA)}, pages 18458--18464. IEEE, 2024.

\bibitem[Ding et~al.(2023)Ding, Rehder, Schneider, Cordts, and Gall]{ding20233dmotformer}
Shuxiao Ding, Eike Rehder, Lukas Schneider, Marius Cordts, and Juergen Gall.
\newblock 3dmotformer: Graph transformer for online 3d multi-object tracking.
\newblock In \emph{Proceedings of the IEEE/CVF international conference on computer vision}, pages 9784--9794, 2023.

\bibitem[Ding et~al.(2024)Ding, Schneider, Cordts, and Gall]{ding2024ada}
Shuxiao Ding, Lukas Schneider, Marius Cordts, and Juergen Gall.
\newblock Ada-track: End-to-end multi-camera 3d multi-object tracking with alternating detection and association.
\newblock In \emph{Proceedings of the IEEE/CVF Conference on Computer Vision and Pattern Recognition}, pages 15184--15194, 2024.

\bibitem[Doll et~al.(2023)Doll, Hanselmann, Schneider, Schulz, Enzweiler, and Lensch]{doll2023star}
Simon Doll, Niklas Hanselmann, Lukas Schneider, Richard Schulz, Markus Enzweiler, and Hendrik~PA Lensch.
\newblock Star-track: Latent motion models for end-to-end 3d object tracking with adaptive spatio-temporal appearance representations.
\newblock \emph{IEEE Robotics and Automation Letters}, 9\penalty0 (2):\penalty0 1326--1333, 2023.

\bibitem[Fan et~al.(2024)Fan, Yu, Yang, Yuan, and Nie]{fan2024quest}
Siqi Fan, Haibao Yu, Wenxian Yang, Jirui Yuan, and Zaiqing Nie.
\newblock Quest: Query stream for practical cooperative perception.
\newblock In \emph{2024 IEEE International Conference on Robotics and Automation (ICRA)}, pages 18436--18442. IEEE, 2024.

\bibitem[Hao et~al.(2024)Hao, Fan, Dai, Zhang, Li, Wang, Yu, Yang, Yuan, and Nie]{hao2024rcooper}
Ruiyang Hao, Siqi Fan, Yingru Dai, Zhenlin Zhang, Chenxi Li, Yuntian Wang, Haibao Yu, Wenxian Yang, Jirui Yuan, and Zaiqing Nie.
\newblock Rcooper: A real-world large-scale dataset for roadside cooperative perception.
\newblock In \emph{Proceedings of the IEEE/CVF Conference on Computer Vision and Pattern Recognition}, pages 22347--22357, 2024.

\bibitem[Hao et~al.(2025)Hao, Jing, Yu, and Nie]{hao2025styledrive}
Ruiyang Hao, Bowen Jing, Haibao Yu, and Zaiqing Nie.
\newblock Styledrive: Towards driving-style aware benchmarking of end-to-end autonomous driving.
\newblock \emph{arXiv preprint arXiv:2506.23982}, 2025.

\bibitem[He et~al.(2016)He, Zhang, Ren, and Sun]{he2016deep}
Kaiming He, Xiangyu Zhang, Shaoqing Ren, and Jian Sun.
\newblock Deep residual learning for image recognition.
\newblock In \emph{Proceedings of the IEEE conference on computer vision and pattern recognition}, pages 770--778, 2016.

\bibitem[Hu et~al.(2022)Hu, Fang, Lei, Zhong, and Chen]{hu2022where2comm}
Yue Hu, Shaoheng Fang, Zixing Lei, Yiqi Zhong, and Siheng Chen.
\newblock Where2comm: Communication-efficient collaborative perception via spatial confidence maps.
\newblock \emph{Advances in neural information processing systems}, 35:\penalty0 4874--4886, 2022.

\bibitem[Hu et~al.(2023)Hu, Lu, Xu, Xie, Chen, and Wang]{hu2023collaboration}
Yue Hu, Yifan Lu, Runsheng Xu, Weidi Xie, Siheng Chen, and Yanfeng Wang.
\newblock Collaboration helps camera overtake lidar in 3d detection.
\newblock In \emph{Proceedings of the IEEE/CVF Conference on Computer Vision and Pattern Recognition}, pages 9243--9252, 2023.

\bibitem[Hu et~al.(2024{\natexlab{a}})Hu, Pang, Qin, Eldar, Chen, Zhang, and Zhang]{hu2024pragmatic}
Yue Hu, Xianghe Pang, Xiaoqi Qin, Yonina~C Eldar, Siheng Chen, Ping Zhang, and Wenjun Zhang.
\newblock Pragmatic communication in multi-agent collaborative perception.
\newblock \emph{arXiv preprint arXiv:2401.12694}, 2024{\natexlab{a}}.

\bibitem[Hu et~al.(2024{\natexlab{b}})Hu, Peng, Liu, Ge, Liu, and Chen]{hu2024communication}
Yue Hu, Juntong Peng, Sifei Liu, Junhao Ge, Si Liu, and Siheng Chen.
\newblock Communication-efficient collaborative perception via information filling with codebook.
\newblock In \emph{Proceedings of the IEEE/CVF Conference on Computer Vision and Pattern Recognition}, pages 15481--15490, 2024{\natexlab{b}}.

\bibitem[Kuhn(1955)]{kuhn1955hungarian}
Harold~W Kuhn.
\newblock The hungarian method for the assignment problem.
\newblock \emph{Naval research logistics quarterly}, 2\penalty0 (1-2):\penalty0 83--97, 1955.

\bibitem[Li et~al.(2024{\natexlab{a}})Li, Zhao, Zhong, Wang, Sun, and Sun]{li2024delving}
Haoyu Li, Yueran Zhao, Jiaru Zhong, Bo Wang, Chao Sun, and Fengchun Sun.
\newblock Delving into the secrets of bev 3d object detection in autonomous driving: A comprehensive survey.
\newblock \emph{Authorea Preprints}, 2024{\natexlab{a}}.

\bibitem[Li and Jin(2022)]{li2022time3d}
Peixuan Li and Jieyu Jin.
\newblock Time3d: End-to-end joint monocular 3d object detection and tracking for autonomous driving.
\newblock In \emph{Proceedings of the IEEE/CVF conference on computer vision and pattern recognition}, pages 3885--3894, 2022.

\bibitem[Li et~al.(2021)Li, Ren, Wu, Chen, Feng, and Zhang]{li2021learning}
Yiming Li, Shunli Ren, Pengxiang Wu, Siheng Chen, Chen Feng, and Wenjun Zhang.
\newblock Learning distilled collaboration graph for multi-agent perception.
\newblock \emph{Advances in Neural Information Processing Systems}, 34:\penalty0 29541--29552, 2021.

\bibitem[Li et~al.(2022)Li, Ma, An, Wang, Zhong, Chen, and Feng]{li2022v2x}
Yiming Li, Dekun Ma, Ziyan An, Zixun Wang, Yiqi Zhong, Siheng Chen, and Chen Feng.
\newblock V2x-sim: Multi-agent collaborative perception dataset and benchmark for autonomous driving.
\newblock \emph{IEEE Robotics and Automation Letters}, 7\penalty0 (4):\penalty0 10914--10921, 2022.

\bibitem[Li et~al.(2023)Li, Yu, Philion, Anandkumar, Fidler, Jia, and Alvarez]{li2023end}
Yanwei Li, Zhiding Yu, Jonah Philion, Anima Anandkumar, Sanja Fidler, Jiaya Jia, and Jose Alvarez.
\newblock End-to-end 3d tracking with decoupled queries.
\newblock In \emph{Proceedings of the IEEE/CVF International Conference on Computer Vision}, pages 18302--18311, 2023.

\bibitem[Li et~al.(2024{\natexlab{b}})Li, Wang, Li, Xie, Sima, Lu, Yu, and Dai]{li2024bevformer}
Zhiqi Li, Wenhai Wang, Hongyang Li, Enze Xie, Chonghao Sima, Tong Lu, Qiao Yu, and Jifeng Dai.
\newblock Bevformer: learning bird's-eye-view representation from lidar-camera via spatiotemporal transformers.
\newblock \emph{IEEE Transactions on Pattern Analysis and Machine Intelligence}, 2024{\natexlab{b}}.

\bibitem[Lin et~al.(2024)Lin, Tian, Duan, Zhou, Zhao, and Cao]{lin2024v2vformer}
Chunmian Lin, Daxin Tian, Xuting Duan, Jianshan Zhou, Dezong Zhao, and Dongpu Cao.
\newblock V2vformer: Vehicle-to-vehicle cooperative perception with spatial-channel transformer.
\newblock \emph{IEEE Transactions on Intelligent Vehicles}, 9\penalty0 (2):\penalty0 3384--3395, 2024.

\bibitem[Lin et~al.(2023)Lin, Pei, Lin, Huang, and Su]{lin2023sparse4d}
Xuewu Lin, Zixiang Pei, Tianwei Lin, Lichao Huang, and Zhizhong Su.
\newblock Sparse4d v3: Advancing end-to-end 3d detection and tracking.
\newblock \emph{arXiv preprint arXiv:2311.11722}, 2023.

\bibitem[Liu et~al.(2025)Liu, Zhong, and Sun]{liu2025bevmamba}
Xiao Liu, Jiaru Zhong, and Chao Sun.
\newblock Bevmamba: Time sequence dense bird’s-eye-view perception modeling with state space model.
\newblock \emph{IEEE Transactions on Intelligent Transportation Systems}, 2025.

\bibitem[Liu et~al.(2022)Liu, Mao, Wu, Feichtenhofer, Darrell, and Xie]{liu2022convnet}
Zhuang Liu, Hanzi Mao, Chao-Yuan Wu, Christoph Feichtenhofer, Trevor Darrell, and Saining Xie.
\newblock A convnet for the 2020s.
\newblock In \emph{Proceedings of the IEEE/CVF conference on computer vision and pattern recognition}, pages 11976--11986, 2022.

\bibitem[Lu et~al.(2023)Lu, Li, Liu, Dianati, Feng, Chen, and Wang]{lu2023robust}
Yifan Lu, Quanhao Li, Baoan Liu, Mehrdad Dianati, Chen Feng, Siheng Chen, and Yanfeng Wang.
\newblock Robust collaborative 3d object detection in presence of pose errors.
\newblock In \emph{2023 IEEE International Conference on Robotics and Automation (ICRA)}, pages 4812--4818. IEEE, 2023.

\bibitem[Luiten et~al.(2020)Luiten, Fischer, and Leibe]{luiten2020track}
Jonathon Luiten, Tobias Fischer, and Bastian Leibe.
\newblock Track to reconstruct and reconstruct to track.
\newblock \emph{IEEE Robotics and Automation Letters}, 5\penalty0 (2):\penalty0 1803--1810, 2020.

\bibitem[Ma et~al.(2024)Ma, Qiao, Zhu, Liu, Kong, Li, Zhou, Kan, and Wu]{ma2024holovic}
Cong Ma, Lei Qiao, Chengkai Zhu, Kai Liu, Zelong Kong, Qing Li, Xueqi Zhou, Yuheng Kan, and Wei Wu.
\newblock Holovic: Large-scale dataset and benchmark for multi-sensor holographic intersection and vehicle-infrastructure cooperative.
\newblock In \emph{Proceedings of the IEEE/CVF Conference on Computer Vision and Pattern Recognition}, pages 22129--22138, 2024.

\bibitem[Mao et~al.(2022)Mao, Guo, Jia, Sun, Zhou, and Niu]{mao2022dolphins}
Ruiqing Mao, Jingyu Guo, Yukuan Jia, Yuxuan Sun, Sheng Zhou, and Zhisheng Niu.
\newblock Dolphins: Dataset for collaborative perception enabled harmonious and interconnected self-driving.
\newblock In \emph{Proceedings of the Asian Conference on Computer Vision}, pages 4361--4377, 2022.

\bibitem[Marinello et~al.(2022)Marinello, Proesmans, and Van~Gool]{marinello2022triplettrack}
Nicola Marinello, Marc Proesmans, and Luc Van~Gool.
\newblock Triplettrack: 3d object tracking using triplet embeddings and lstm.
\newblock In \emph{Proceedings of the IEEE/CVF conference on computer vision and pattern recognition}, pages 4500--4510, 2022.

\bibitem[Pang et~al.(2022)Pang, Li, and Wang]{pang2022simpletrack}
Ziqi Pang, Zhichao Li, and Naiyan Wang.
\newblock Simpletrack: Understanding and rethinking 3d multi-object tracking.
\newblock In \emph{European conference on computer vision}, pages 680--696. Springer, 2022.

\bibitem[Pang et~al.(2023)Pang, Li, Tokmakov, Chen, Zagoruyko, and Wang]{pang2023standing}
Ziqi Pang, Jie Li, Pavel Tokmakov, Dian Chen, Sergey Zagoruyko, and Yu-Xiong Wang.
\newblock Standing between past and future: Spatio-temporal modeling for multi-camera 3d multi-object tracking.
\newblock In \emph{Proceedings of the IEEE/CVF conference on computer vision and pattern recognition}, pages 17928--17938, 2023.

\bibitem[Park et~al.(2022)Park, Xu, Yang, Keutzer, Kitani, Tomizuka, and Zhan]{park2022time}
Jinhyung Park, Chenfeng Xu, Shijia Yang, Kurt Keutzer, Kris Kitani, Masayoshi Tomizuka, and Wei Zhan.
\newblock Time will tell: New outlooks and a baseline for temporal multi-view 3d object detection.
\newblock \emph{arXiv preprint arXiv:2210.02443}, 2022.

\bibitem[Qi et~al.(2017)Qi, Su, Mo, and Guibas]{qi2017pointnet}
Charles~R Qi, Hao Su, Kaichun Mo, and Leonidas~J Guibas.
\newblock Pointnet: Deep learning on point sets for 3d classification and segmentation.
\newblock In \emph{Proceedings of the IEEE conference on computer vision and pattern recognition}, pages 652--660, 2017.

\bibitem[Ruan et~al.(2025)Ruan, Yu, Yang, Fan, and Nie]{ruan2025learning}
Hongzhi Ruan, Haibao Yu, Wenxian Yang, Siqi Fan, and Zaiqing Nie.
\newblock Learning cooperative trajectory representations for motion forecasting.
\newblock \emph{Advances in Neural Information Processing Systems}, 37:\penalty0 13430--13457, 2025.

\bibitem[Ruppel et~al.(2022)Ruppel, Faion, Gl{\"a}ser, and Dietmayer]{ruppel2022transformers}
Felicia Ruppel, Florian Faion, Claudius Gl{\"a}ser, and Klaus Dietmayer.
\newblock Transformers for multi-object tracking on point clouds.
\newblock In \emph{2022 IEEE Intelligent Vehicles Symposium (IV)}, pages 852--859. IEEE, 2022.

\bibitem[Sadjadpour et~al.(2023)Sadjadpour, Li, Ambrus, and Bohg]{sadjadpour2023shasta}
Tara Sadjadpour, Jie Li, Rares Ambrus, and Jeannette Bohg.
\newblock Shasta: Modeling shape and spatio-temporal affinities for 3d multi-object tracking.
\newblock \emph{IEEE Robotics and Automation Letters}, 9\penalty0 (5):\penalty0 4273--4280, 2023.

\bibitem[Song et~al.(2024)Song, Liang, Cao, Yan, Zimmer, Gross, Festag, and Knoll]{song2024collaborative}
Rui Song, Chenwei Liang, Hu Cao, Zhiran Yan, Walter Zimmer, Markus Gross, Andreas Festag, and Alois Knoll.
\newblock Collaborative semantic occupancy prediction with hybrid feature fusion in connected automated vehicles.
\newblock In \emph{Proceedings of the IEEE/CVF Conference on Computer Vision and Pattern Recognition}, pages 17996--18006, 2024.

\bibitem[Stearns et~al.(2022)Stearns, Rempe, Li, Ambru{\c{s}}, Zakharov, Guizilini, Yang, and Guibas]{stearns2022spot}
Colton Stearns, Davis Rempe, Jie Li, Rare{\c{s}} Ambru{\c{s}}, Sergey Zakharov, Vitor Guizilini, Yanchao Yang, and Leonidas~J Guibas.
\newblock Spot: Spatiotemporal modeling for 3d object tracking.
\newblock In \emph{European Conference on Computer Vision}, pages 639--656. Springer, 2022.

\bibitem[Su et~al.(2023)Su, Arakawa, and Murata]{su20233d}
Hao Su, Shin’Ichi Arakawa, and Masayuki Murata.
\newblock 3d multi-object tracking based on two-stage data association for collaborative perception scenarios.
\newblock In \emph{2023 IEEE Intelligent Vehicles Symposium (IV)}, pages 1--7. IEEE, 2023.

\bibitem[Su et~al.(2024{\natexlab{a}})Su, Arakawa, and Murata]{su2024cooperative}
Hao Su, Shin’ichi Arakawa, and Masayuki Murata.
\newblock Cooperative 3d multi-object tracking for connected and automated vehicles with complementary data association.
\newblock In \emph{2024 IEEE Intelligent Vehicles Symposium (IV)}, pages 285--291. IEEE, 2024{\natexlab{a}}.

\bibitem[Su et~al.(2024{\natexlab{b}})Su, Han, Li, Zhang, Feng, Ding, and Miao]{su2024collaborative}
Sanbao Su, Songyang Han, Yiming Li, Zhili Zhang, Chen Feng, Caiwen Ding, and Fei Miao.
\newblock Collaborative multi-object tracking with conformal uncertainty propagation.
\newblock \emph{IEEE Robotics and Automation Letters}, 9\penalty0 (4):\penalty0 3323--3330, 2024{\natexlab{b}}.

\bibitem[Vaswani et~al.(2017)Vaswani, Shazeer, Parmar, Uszkoreit, Jones, Gomez, Kaiser, and Polosukhin]{vaswani2017attention}
Ashish Vaswani, Noam Shazeer, Niki Parmar, Jakob Uszkoreit, Llion Jones, Aidan~N Gomez, {\L}ukasz Kaiser, and Illia Polosukhin.
\newblock Attention is all you need.
\newblock \emph{Advances in neural information processing systems}, 30, 2017.

\bibitem[Wang et~al.(2025)Wang, Cao, Zhong, Zhang, Yu, He, and Xu]{wang2025griffin}
Jiahao Wang, Xiangyu Cao, Jiaru Zhong, Yuner Zhang, Haibao Yu, Lei He, and Shaobing Xu.
\newblock Griffin: Aerial-ground cooperative detection and tracking dataset and benchmark.
\newblock \emph{arXiv preprint arXiv:2503.06983}, 2025.

\bibitem[Wang et~al.(2023{\natexlab{a}})Wang, Zhang, Qin, Li, Gao, Yang, Li, Li, Zhu, Wang, et~al.]{wang2023camo}
Li Wang, Xinyu Zhang, Wenyuan Qin, Xiaoyu Li, Jinghan Gao, Lei Yang, Zhiwei Li, Jun Li, Lei Zhu, Hong Wang, et~al.
\newblock Camo-mot: Combined appearance-motion optimization for 3d multi-object tracking with camera-lidar fusion.
\newblock \emph{IEEE Transactions on Intelligent Transportation Systems}, 24\penalty0 (11):\penalty0 11981--11996, 2023{\natexlab{a}}.

\bibitem[Wang et~al.(2024{\natexlab{a}})Wang, He, Chen, and Zhang]{wang2024onetrack}
Qitai Wang, Jiawei He, Yuntao Chen, and Zhaoxiang Zhang.
\newblock Onetrack: Demystifying the conflict between detection and tracking in end-to-end 3d trackers.
\newblock In \emph{European Conference on Computer Vision}, pages 387--404. Springer, 2024{\natexlab{a}}.

\bibitem[Wang et~al.(2023{\natexlab{b}})Wang, Liu, Wang, Li, and Zhang]{wang2023exploring}
Shihao Wang, Yingfei Liu, Tiancai Wang, Ying Li, and Xiangyu Zhang.
\newblock Exploring object-centric temporal modeling for efficient multi-view 3d object detection.
\newblock In \emph{Proceedings of the IEEE/CVF international conference on computer vision}, pages 3621--3631, 2023{\natexlab{b}}.

\bibitem[Wang et~al.(2024{\natexlab{b}})Wang, Fu, He, Huang, Meng, Zhang, Zhou, Xu, and Zhang]{wang2024multi}
Xiyang Wang, Chunyun Fu, Jiawei He, Mingguang Huang, Ting Meng, Siyu Zhang, Hangning Zhou, Ziyao Xu, and Chi Zhang.
\newblock A multi-modal fusion-based 3d multi-object tracking framework with joint detection.
\newblock \emph{IEEE Robotics and Automation Letters}, 2024{\natexlab{b}}.

\bibitem[Weng et~al.(2020{\natexlab{a}})Weng, Wang, Held, and Kitani]{weng20203d}
Xinshuo Weng, Jianren Wang, David Held, and Kris Kitani.
\newblock 3d multi-object tracking: A baseline and new evaluation metrics.
\newblock In \emph{2020 IEEE/RSJ International Conference on Intelligent Robots and Systems (IROS)}, pages 10359--10366. IEEE, 2020{\natexlab{a}}.

\bibitem[Weng et~al.(2020{\natexlab{b}})Weng, Wang, Man, and Kitani]{weng2020gnn3dmot}
Xinshuo Weng, Yongxin Wang, Yunze Man, and Kris~M Kitani.
\newblock Gnn3dmot: Graph neural network for 3d multi-object tracking with 2d-3d multi-feature learning.
\newblock In \emph{Proceedings of the IEEE/CVF conference on computer vision and pattern recognition}, pages 6499--6508, 2020{\natexlab{b}}.

\bibitem[Xiang et~al.(2024)Xiang, Zheng, Xia, Xu, Gao, Zhou, Han, Ji, Li, Meng, et~al.]{xiang2024v2x}
Hao Xiang, Zhaoliang Zheng, Xin Xia, Runsheng Xu, Letian Gao, Zewei Zhou, Xu Han, Xinkai Ji, Mingxi Li, Zonglin Meng, et~al.
\newblock V2x-real: a largs-scale dataset for vehicle-to-everything cooperative perception.
\newblock In \emph{European Conference on Computer Vision}, pages 455--470. Springer, 2024.

\bibitem[Xu et~al.(2023{\natexlab{a}})Xu, Zhang, Lin, Qian, and He]{xu2023deformable}
Baixin Xu, Jiarui Zhang, Kwan-Yee Lin, Chen Qian, and Ying He.
\newblock Deformable model-driven neural rendering for high-fidelity 3d reconstruction of human heads under low-view settings.
\newblock In \emph{Proceedings of the IEEE/CVF International Conference on Computer Vision}, pages 17924--17934, 2023{\natexlab{a}}.

\bibitem[Xu et~al.(2024)Xu, Hu, Hou, Lin, Wu, Qian, and He]{xu2024parameterization}
Baixin Xu, Jiangbei Hu, Fei Hou, Kwan-Yee Lin, Wayne Wu, Chen Qian, and Ying He.
\newblock Parameterization-driven neural surface reconstruction for object-oriented editing in neural rendering.
\newblock In \emph{European Conference on Computer Vision}, pages 461--479. Springer, 2024.

\bibitem[Xu et~al.(2025)Xu, Shao, Wang, Liu, Yang, Li, and Wang]{xu2025component}
Jiahui Xu, Wenbo Shao, Weida Wang, Cheng Liu, Chao Yang, Jun Li, and Hong Wang.
\newblock From component to system: A task-unified planning system with planning-oriented predictor.
\newblock \emph{IEEE Transactions on Vehicular Technology}, 74\penalty0 (4):\penalty0 5335--5348, 2025.

\bibitem[Xu et~al.(2022{\natexlab{a}})Xu, Tu, Xiang, Shao, Zhou, and Ma]{xu2022cobevt}
Runsheng Xu, Zhengzhong Tu, Hao Xiang, Wei Shao, Bolei Zhou, and Jiaqi Ma.
\newblock Cobevt: Cooperative bird's eye view semantic segmentation with sparse transformers.
\newblock \emph{arXiv preprint arXiv:2207.02202}, 2022{\natexlab{a}}.

\bibitem[Xu et~al.(2022{\natexlab{b}})Xu, Xiang, Tu, Xia, Yang, and Ma]{xu2022v2x}
Runsheng Xu, Hao Xiang, Zhengzhong Tu, Xin Xia, Ming-Hsuan Yang, and Jiaqi Ma.
\newblock V2x-vit: Vehicle-to-everything cooperative perception with vision transformer.
\newblock In \emph{European conference on computer vision}, pages 107--124. Springer, 2022{\natexlab{b}}.

\bibitem[Xu et~al.(2022{\natexlab{c}})Xu, Xiang, Xia, Han, Li, and Ma]{xu2022opv2v}
Runsheng Xu, Hao Xiang, Xin Xia, Xu Han, Jinlong Li, and Jiaqi Ma.
\newblock Opv2v: An open benchmark dataset and fusion pipeline for perception with vehicle-to-vehicle communication.
\newblock In \emph{2022 International Conference on Robotics and Automation (ICRA)}, pages 2583--2589. IEEE, 2022{\natexlab{c}}.

\bibitem[Xu et~al.(2023{\natexlab{b}})Xu, Xia, Li, Li, Zhang, Tu, Meng, Xiang, Dong, Song, et~al.]{xu2023v2v4real}
Runsheng Xu, Xin Xia, Jinlong Li, Hanzhao Li, Shuo Zhang, Zhengzhong Tu, Zonglin Meng, Hao Xiang, Xiaoyu Dong, Rui Song, et~al.
\newblock V2v4real: A real-world large-scale dataset for vehicle-to-vehicle cooperative perception.
\newblock In \emph{Proceedings of the IEEE/CVF Conference on Computer Vision and Pattern Recognition}, pages 13712--13722, 2023{\natexlab{b}}.

\bibitem[Yan et~al.(2024)Yan, Liu, Ai, Li, Wan, and Pu]{yan2024pointssc}
Yuxiang Yan, Boda Liu, Jianfei Ai, Qinbu Li, Ru Wan, and Jian Pu.
\newblock Pointssc: A cooperative vehicle-infrastructure point cloud benchmark for semantic scene completion.
\newblock In \emph{2024 IEEE International Conference on Robotics and Automation (ICRA)}, pages 17027--17034. IEEE, 2024.

\bibitem[Yang et~al.(2022)Yang, Yu, Li, Li, and Tao]{yang2022quality}
Jinrong Yang, En Yu, Zeming Li, Xiaoping Li, and Wenbing Tao.
\newblock Quality matters: Embracing quality clues for robust 3d multi-object tracking.
\newblock \emph{arXiv preprint arXiv:2208.10976}, 2022.

\bibitem[Yang et~al.(2023)Yang, Yang, Zhang, Wang, Sun, and Song]{yang2023what2comm}
Kun Yang, Dingkang Yang, Jingyu Zhang, Hanqi Wang, Peng Sun, and Liang Song.
\newblock What2comm: Towards communication-efficient collaborative perception via feature decoupling.
\newblock In \emph{Proceedings of the 31st ACM international conference on multimedia}, pages 7686--7695, 2023.

\bibitem[Yang et~al.(2024{\natexlab{a}})Yang, Zhang, Li, Wang, Song, Zhao, Song, Wang, Zhou, Shen, et~al.]{yang2024v2x}
Lei Yang, Xinyu Zhang, Jun Li, Chen Wang, Zhiying Song, Tong Zhao, Ziying Song, Li Wang, Mo Zhou, Yang Shen, et~al.
\newblock V2x-radar: A multi-modal dataset with 4d radar for cooperative perception.
\newblock \emph{arXiv preprint arXiv:2411.10962}, 2024{\natexlab{a}}.

\bibitem[Yang et~al.(2024{\natexlab{b}})Yang, Mao, Yang, Ai, Kong, Yu, and Zhang]{yang2024lidar}
Zhenwei Yang, Jilei Mao, Wenxian Yang, Yibo Ai, Yu Kong, Haibao Yu, and Weidong Zhang.
\newblock Lidar-based end-to-end temporal perception for vehicle-infrastructure cooperation.
\newblock \emph{arXiv preprint arXiv:2411.14927}, 2024{\natexlab{b}}.

\bibitem[Yin et~al.(2025)Yin, Xu, Li, Zhang, and Rigoll]{yin2025knowledge}
Huilin Yin, Yangwenhui Xu, Jiaxiang Li, Hao Zhang, and Gerhard Rigoll.
\newblock Knowledge-informed multi-agent trajectory prediction at signalized intersections for infrastructure-to-everything.
\newblock \emph{arXiv preprint arXiv:2501.13461}, 2025.

\bibitem[Yin et~al.(2021)Yin, Zhou, and Krahenbuhl]{yin2021center}
Tianwei Yin, Xingyi Zhou, and Philipp Krahenbuhl.
\newblock Center-based 3d object detection and tracking.
\newblock In \emph{Proceedings of the IEEE/CVF conference on computer vision and pattern recognition}, pages 11784--11793, 2021.

\bibitem[Yu et~al.(2022)Yu, Luo, Shu, Huo, Yang, Shi, Guo, Li, Hu, Yuan, et~al.]{yu2022dair}
Haibao Yu, Yizhen Luo, Mao Shu, Yiyi Huo, Zebang Yang, Yifeng Shi, Zhenglong Guo, Hanyu Li, Xing Hu, Jirui Yuan, et~al.
\newblock Dair-v2x: A large-scale dataset for vehicle-infrastructure cooperative 3d object detection.
\newblock In \emph{Proceedings of the IEEE/CVF Conference on Computer Vision and Pattern Recognition}, pages 21361--21370, 2022.

\bibitem[Yu et~al.(2023{\natexlab{a}})Yu, Tang, Xie, Mao, Luo, and Nie]{yu2023flow}
Haibao Yu, Yingjuan Tang, Enze Xie, Jilei Mao, Ping Luo, and Zaiqing Nie.
\newblock Flow-based feature fusion for vehicle-infrastructure cooperative 3d object detection.
\newblock \emph{Advances in Neural Information Processing Systems}, 36:\penalty0 34493--34503, 2023{\natexlab{a}}.

\bibitem[Yu et~al.(2023{\natexlab{b}})Yu, Yang, Ruan, Yang, Tang, Gao, Hao, Shi, Pan, Sun, et~al.]{yu2023v2x}
Haibao Yu, Wenxian Yang, Hongzhi Ruan, Zhenwei Yang, Yingjuan Tang, Xu Gao, Xin Hao, Yifeng Shi, Yifeng Pan, Ning Sun, et~al.
\newblock V2x-seq: A large-scale sequential dataset for vehicle-infrastructure cooperative perception and forecasting.
\newblock In \emph{Proceedings of the IEEE/CVF Conference on Computer Vision and Pattern Recognition}, pages 5486--5495, 2023{\natexlab{b}}.

\bibitem[Yu et~al.(2025)Yu, Yang, Zhong, Yang, Fan, Luo, and Nie]{yu2024end}
Haibao Yu, Wenxian Yang, Jiaru Zhong, Zhenwei Yang, Siqi Fan, Ping Luo, and Zaiqing Nie.
\newblock End-to-end autonomous driving through v2x cooperation.
\newblock In \emph{The 39th Annual AAAI Conference on Artificial Intelligence}, 2025.

\bibitem[Yue et~al.(2024)Yue, Zhong, Ning, Chen, Sun, Wang, Lian, Li, and Sun]{yue2024development}
Chao Yue, Jiaru Zhong, Qili Ning, Xiaohui Chen, Chao Sun, Wenwei Wang, Yubo Lian, Keqiang Li, and Fengchun Sun.
\newblock Development strategy of vehicle-energy-road-cloud collaboration in the new technological situation.
\newblock \emph{Strategic Study of Chinese Academy of Engineering}, 26\penalty0 (1):\penalty0 45--58, 2024.

\bibitem[Zeng et~al.(2022)Zeng, Dong, Zhang, Wang, Zhang, and Wei]{zeng2022motr}
Fangao Zeng, Bin Dong, Yuang Zhang, Tiancai Wang, Xiangyu Zhang, and Yichen Wei.
\newblock Motr: End-to-end multiple-object tracking with transformer.
\newblock In \emph{European conference on computer vision}, pages 659--675. Springer, 2022.

\bibitem[Zhang et~al.(2022)Zhang, Chen, Wang, Wang, and Zhao]{zhang2022mutr3d}
Tianyuan Zhang, Xuanyao Chen, Yue Wang, Yilun Wang, and Hang Zhao.
\newblock Mutr3d: A multi-camera tracking framework via 3d-to-2d queries.
\newblock In \emph{Proceedings of the IEEE/CVF Conference on Computer Vision and Pattern Recognition}, pages 4537--4546, 2022.

\bibitem[Zhang et~al.(2025)Zhang, Zhou, Wang, Ji, Huang, and Chen]{zhang2025co}
Xinyu Zhang, Zewei Zhou, Zhaoyi Wang, Yangjie Ji, Yanjun Huang, and Hong Chen.
\newblock Co-mtp: A cooperative trajectory prediction framework with multi-temporal fusion for autonomous driving.
\newblock \emph{arXiv preprint arXiv:2502.16589}, 2025.

\bibitem[Zhong et~al.(2024)Zhong, Yu, Zhu, Xu, Yang, Nie, and Sun]{zhong2024leveraging}
Jiaru Zhong, Haibao Yu, Tianyi Zhu, Jiahui Xu, Wenxian Yang, Zaiqing Nie, and Chao Sun.
\newblock Leveraging temporal contexts to enhance vehicle-infrastructure cooperative perception.
\newblock In \emph{2024 IEEE 27th International Conference on Intelligent Transportation Systems (ITSC)}. IEEE, 2024.

\bibitem[Zhou et~al.(2024{\natexlab{a}})Zhou, Tang, Hao, He, Ho, Gu, Hou, Hao, Sun, Zhan, et~al.]{zhou2024ua}
Lijun Zhou, Tao Tang, Pengkun Hao, Zihang He, Kalok Ho, Shuo Gu, Wenbo Hou, Zhihui Hao, Haiyang Sun, Kun Zhan, et~al.
\newblock Ua-track: Uncertainty-aware end-to-end 3d multi-object tracking.
\newblock \emph{arXiv preprint arXiv:2406.02147}, 2024{\natexlab{a}}.

\bibitem[Zhou et~al.(2019)Zhou, Barnes, Lu, Yang, and Li]{zhou2019continuity}
Yi Zhou, Connelly Barnes, Jingwan Lu, Jimei Yang, and Hao Li.
\newblock On the continuity of rotation representations in neural networks.
\newblock In \emph{Proceedings of the IEEE/CVF conference on computer vision and pattern recognition}, pages 5745--5753, 2019.

\bibitem[Zhou et~al.(2024{\natexlab{b}})Zhou, Xiang, Zheng, Zhao, Lei, Zhang, Cai, Liu, Liu, Bajji, et~al.]{zhou2024v2xpnp}
Zewei Zhou, Hao Xiang, Zhaoliang Zheng, Seth~Z Zhao, Mingyue Lei, Yun Zhang, Tianhui Cai, Xinyi Liu, Johnson Liu, Maheswari Bajji, et~al.
\newblock V2xpnp: Vehicle-to-everything spatio-temporal fusion for multi-agent perception and prediction.
\newblock \emph{arXiv preprint arXiv:2412.01812}, 2024{\natexlab{b}}.

\bibitem[Zhu et~al.(2024)Zhu, Leng, Zhong, Zhang, and Sun]{zhu2024lanemapnet}
Tianyi Zhu, Jianghao Leng, Jiaru Zhong, Zhang Zhang, and Chao Sun.
\newblock Lanemapnet: Lane network recognization and hd map construction using curve region aware temporal bird’s-eye-view perception.
\newblock In \emph{2024 IEEE Intelligent Vehicles Symposium (IV)}, pages 2168--2175. IEEE, 2024.

\bibitem[Zimmer et~al.(2024)Zimmer, Wardana, Sritharan, Zhou, Song, and Knoll]{zimmer2024tumtraf}
Walter Zimmer, Gerhard~Arya Wardana, Suren Sritharan, Xingcheng Zhou, Rui Song, and Alois~C Knoll.
\newblock Tumtraf v2x cooperative perception dataset.
\newblock In \emph{Proceedings of the IEEE/CVF conference on computer vision and pattern recognition}, pages 22668--22677, 2024.

\end{thebibliography}
}

\clearpage
\appendix
\section{Appendix Overview}
\label{sec:appendix}
In the appendix, we present 1) additional implementation details in \cref{supp:experiment}; 2) extended experimental results in \cref{supp:results}, covering inference speed analysis, backbone architecture comparisons, ablation studies on roadside data contribution, communication latency tests, and pose estimation error impacts; 3) qualitative analyses of instance association and tracking results in \cref{supp:visualization}.

\section{More Details of Experiments}
\label{supp:experiment}
\subsection{Implementation Details}
We use two versions of the backbone: ResNet50 and ResNet101. For ResNet50, we crop the images to \(540 \times 960\) and set the BEV feature size to \(50 \times 50\). For the larger one, we keep the input image size unchanged and set the BEV size to \(200 \times 200\). For Griffin, we only trained the ResNet-50 version.
To reduce memory consumption, we adopt a streaming video training approach \cite{park2022time, wang2023exploring}, which may slow down model convergence. Consequently, for V2X-Seq, the vehicle and infrastructure models are trained for 48 and 24 epochs respectively in the first stage, while the cooperative model undergoes 48 epochs of training.
For Griffin, the vehicle and UAV models are each trained for 48 epochs in the first stage, followed by 48 epochs of training for the second-stage cooperative model.

\subsection{Baseline Settings} To demonstrate the superiority of our method, we compare it with several existing cooperative tracking methods: (1) No Fusion: This baseline uses only the ego vehicle's images as input and does not activate the cooperative module. (2) Late Fusion + AB3DMOT \cite{weng20203d}: This method follows the tracking by cooperative detection paradigm, where detection results from multiple agents are fused, and the cooperative detection results are fed into the classic tracking method AB3DMOT \cite{weng20203d}. For a fair comparison, we use BEVFormer \cite{li2024bevformer} as the detector and implement late fusion following DAIR-V2X \cite{yu2022dair}. (3) BEV Feature Fusion + AB3DMOT \cite{weng20203d}: This method also follows the tracking by cooperative detection paradigm but uses feature fusion for cooperative detection. Specifically, we use BEVFormer \cite{li2024bevformer} as the detector, align the two BEV features spatially based on relative positions, and then fuse the concatenated BEV features using a multi-layer convolution neural network before feeding them into the decoder. (4) UniV2X \cite{yu2024end}: This is the first end-to-end cooperative planning method, which includes modules for tracking, mapping, occupancy, and planning. Since we focus on the 3D MOT task, we retain only the Agent Fusion module and the tracking framework, referred to as UniV2X-Track, which belongs to the end-to-end cooperative tracking paradigm mentioned earlier. (5) Other SOTA methods, including V2X-ViT \cite{xu2022v2x}, where2comm \cite{hu2022where2comm}, DiscoNet \cite{li2021learning}, CoAlign \cite{lu2023robust}. For a fair comparison, they use the same inputs and evaluation settings as ours. All methods, except for CoCa3D \cite{hu2023collaboration} based on depth estimation, are implemented using BEVFormer \cite{li2024bevformer}.

\begin{table}[t]
    \centering
    \setlength{\tabcolsep}{10pt}
    \renewcommand\arraystretch{0.9}
    \scalebox{0.95}
    {
    \begin{tabular}{ccc}
    \toprule
      Backbone & mAP$\uparrow$ & AMOTA$\uparrow$ \\
    \midrule
      ResNet101 &  0.390 & 0.328 \\
    \rowcolor[gray]{.9}
      ConvNeXt-S & 0.413~\textcolor{red}{(+0.023)} & 0.429~\textcolor{red}{(+0.101)} \\
    \bottomrule
    \end{tabular}
    }
    \caption{\textbf{Performance of Different Backbones on the V2X-Seq.}}
    \label{tab:sup_backbone}
\end{table}

\begin{table}[t]
    \centering
    \renewcommand\arraystretch{0.9}
    \setlength{\tabcolsep}{3.0pt}
    \scalebox{0.9}
    {
    \begin{tabular}[]{ccccc}
    \toprule
      Method & MDFE & CAA & GBA+Aggr. & Total \\
      \midrule
      CoopTrack-ResNet50 & 42.01 & 2.08 & 9.95 & 121.88 \\
      CoopTrack-ResNet101 & 131.91 & 2.04 & 8.68 & 207.99 \\
      \bottomrule
    \end{tabular}
    }
    \caption{\textbf{Runtime of Key Modules of CoopTrack} \textit{(unit: ms)}. MDFE stands for the multi-dimensional feature extraction module, CAA for the cross-agent alignment module, GBA for the graph-based association module, and Aggr. for feature aggregation.}
    \label{tab:sup_runtime}
\end{table}

\begin{figure}[t]
    \centering
    \includegraphics[width=1.0\linewidth]{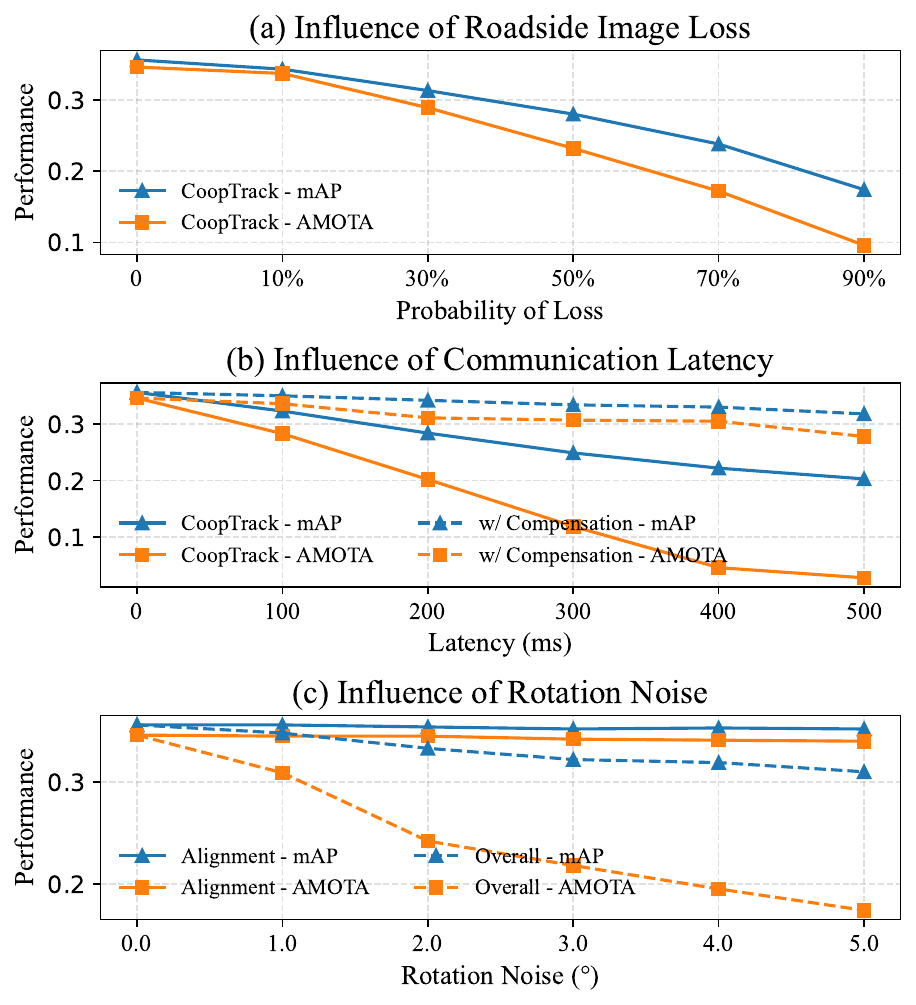}
    \caption{\textbf{Comparison of Performance in Different Conditions.}}
    \label{fig:sup_robust}
\end{figure}

\begin{figure*}[t]
    \centering
    \begin{subfigure}[b]{0.47\textwidth}
        \centering
        \includegraphics[width=\textwidth]{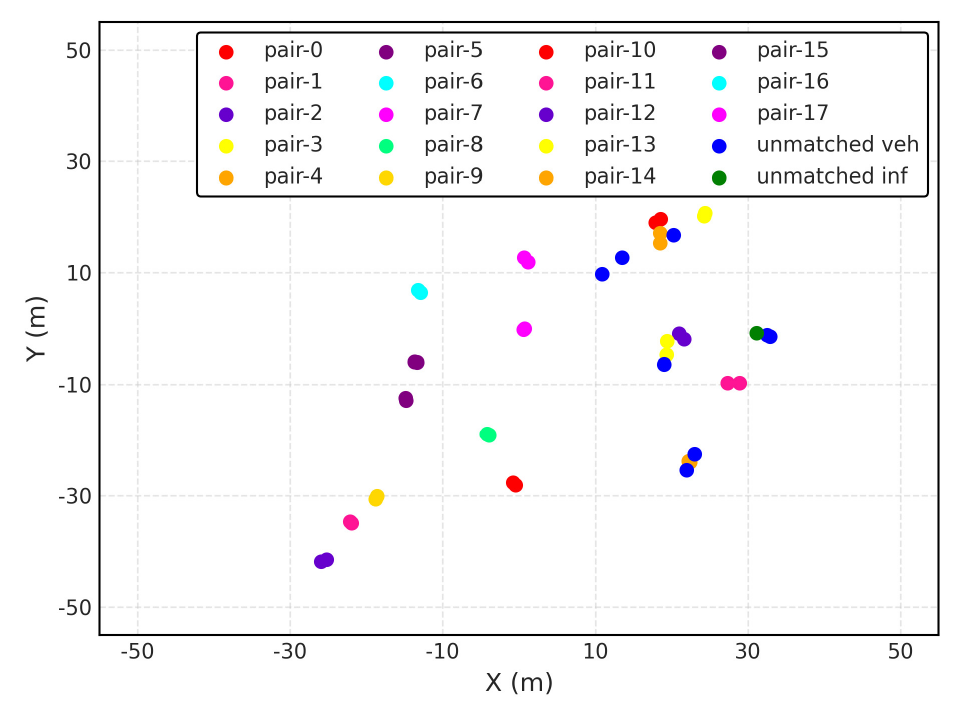}
        \caption{Example Frame 1}
    \end{subfigure}
    \begin{subfigure}[b]{0.47\textwidth}
        \centering
        \includegraphics[width=\textwidth]{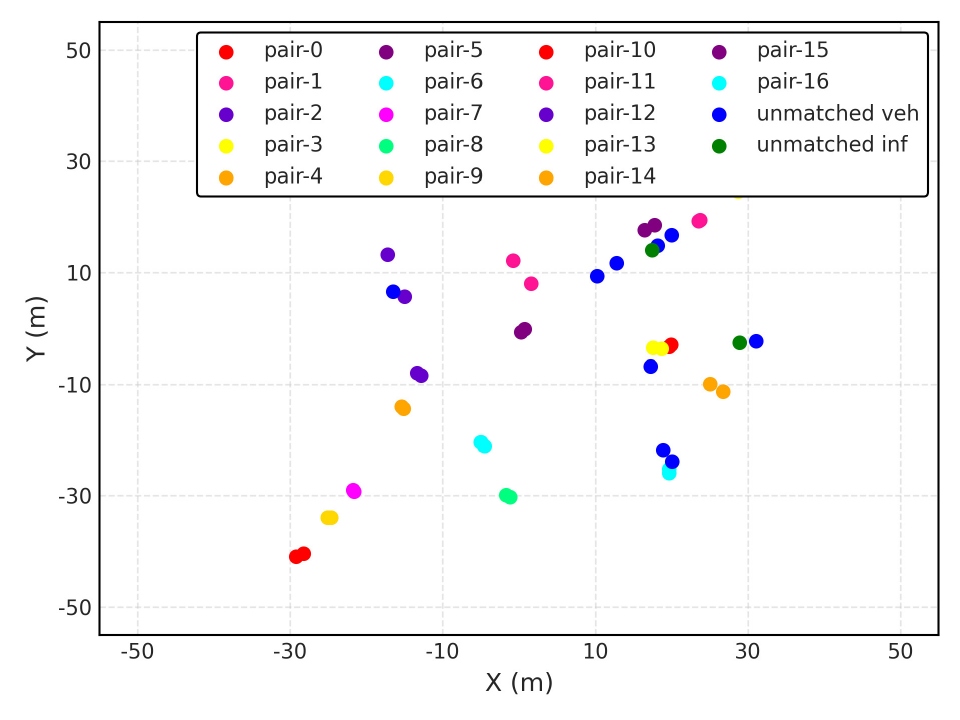}
        \caption{Example Frame 2}
    \end{subfigure}

    \begin{subfigure}[b]{0.47\textwidth}
        \centering
        \includegraphics[width=\textwidth]{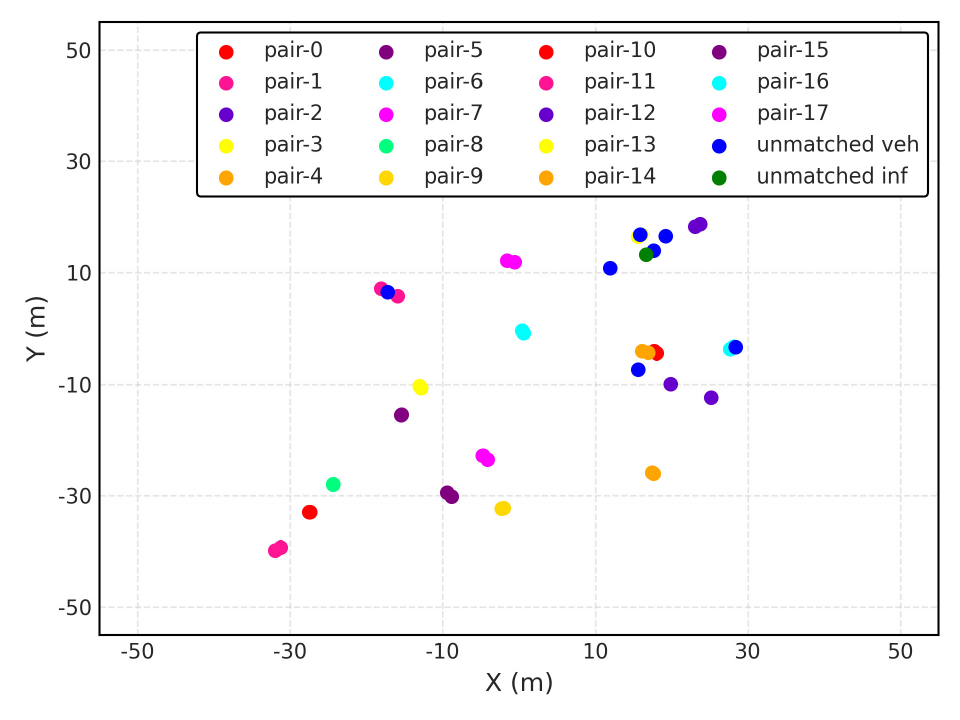}
        \caption{Example Frame 3}
    \end{subfigure}
    \begin{subfigure}[b]{0.47\textwidth}
        \centering
        \includegraphics[width=\textwidth]{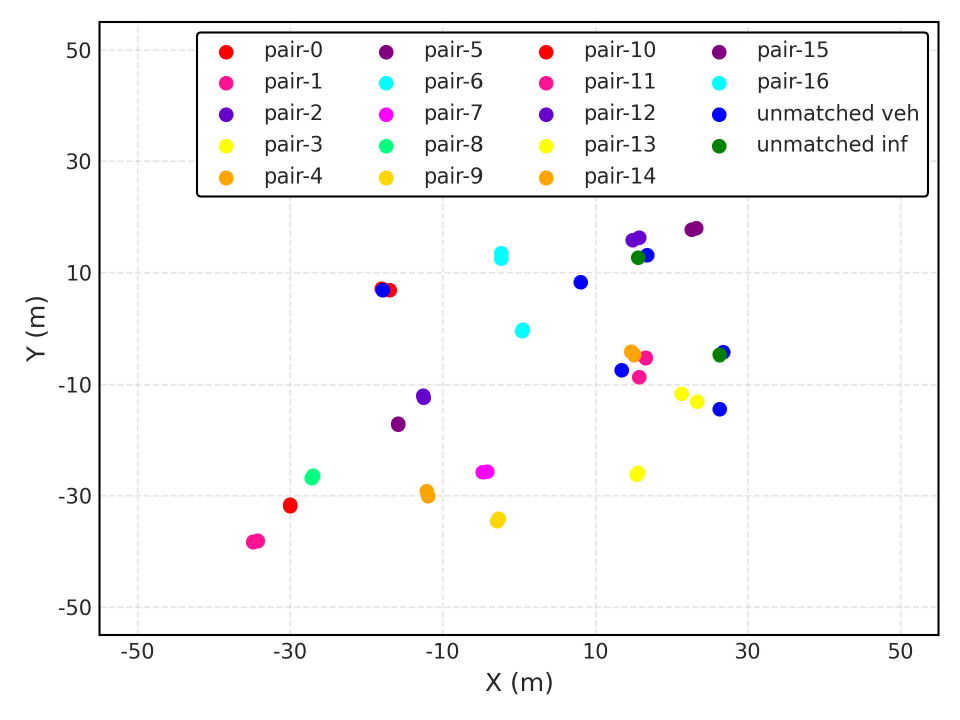}
        \caption{Example Frame 4}
    \end{subfigure}

    \caption{\textbf{Visualization of CoopTrack's association results on the V2X-Seq Dataset.} We visualize the association results of several key frames in a sequence by plotting instances within the ego-vehicle-centric BEV coordinate frame, where each instance is represented by its reference point. Matched instance pairs, including a vehicle instance and an infrastructure instance, are uniquely color-coded according to the ID, while unmatched vehicle and roadside instances are marked distinctly in blue and green, respectively. Note that due to the close proximity of instances, there is overlap in the figure.}
    \label{fig:sup_association_visualization}
\end{figure*}

\subsection{Evaluation Metrics} To assess performance, we utilize widely recognized metrics in 3D object detection and tracking \cite{caesarNuScenesMultimodalDataset2020}, among which the primary indicators are Mean Average Precision (mAP) and Average Multi-Object Tracking Accuracy (AMOTA). Furthermore, to evaluate the transmission costs inherent in cooperative approaches, we employ Bytes per second (BPS) as another essential metric \cite{yu2022dair,yu2023v2x}. To ensure fair comparison with existing methods, we follow the evaluation protocols established by UniV2X \cite{yu2024end} and Griffin \cite{wang2025griffin}, reporting performance metrics exclusively for the vehicle category.

\section{More Experiments}
\label{supp:results}
\subsection{Recent Image Backbone} We upgrade the image feature extraction backbone by replacing ResNet101 \cite{he2016deep} with the more recent ConvNeXt-Small \cite{liu2022convnet} architecture. As demonstrated in our V2X-Seq experiments (see \cref{tab:sup_backbone}), this modification yields significant performance gains of +2.3\% mAP and +10.1\% AMOTA, confirming the scalability of our approach through backbone compatibility.

\subsection{Inference Speed} To evaluate computational efficiency, we measure the average inference time on a single NVIDIA RTX 3090 GPU across the V2X-Seq validation set, with detailed results presented in Table \ref{tab:sup_runtime}. It can be observed that the time consumption is primarily concentrated in the multi-dimensional feature extraction module, while cross-agent alignment and aggregation do not take much time. Consequently, with the ResNet50 backbone, CoopTrack achieves nearly real-time performance at approximately 10Hz.

\subsection{Ablation Study of Infrastructure Images} To investigate CoopTrack's dependence on roadside data, we evaluate the model without infrastructure image inputs. As shown in \cref{fig:sup_robust}(a), while performance degrades without roadside images, the system still surpasses the No Fusion baseline (0.110 mAP and 0.087 AMOTA), demonstrating the inherent robustness of our approach. This suggests that while roadside information enhances perception accuracy, the framework maintains functional capability when operating independently.

\subsection{Influence of Communication Latency} As a critical challenge in real-world cooperative perception systems, communication latency induces spatiotemporal misalignment between cooperative data and ego-vehicle observations, significantly degrading perception performance \cite{yu2023flow}. To analyze its impact on CoopTrack, we simulate delayed infrastructure-to-vehicle communication by introducing artificially lagged roadside data. As shown in \cref{fig:sup_robust}(b), experimental results demonstrate that under 500ms latency, the system exhibits substantial performance degradation of 15.3\% in mAP and 31.8\% in AMOTA, highlighting the importance of latency mitigation for practical deployment.

To mitigate this issue, we introduce the feature flow prediction module \cite{yu2023flow} that leverages historical query states to learn temporal dynamics and calibrate incoming infrastructure queries based on their timestamps. As shown in \cref{fig:sup_robust}(b), this module significantly enhances CoopTrack’s robustness to latency, limiting performance degradation under 500ms delay to just 3.8\% in mAP and 6.8\% in AMOTA, a marked improvement over the uncompensated baseline. While this method reflects substantial progress, further research is needed to develop advanced temporal modeling techniques and adaptive compensation strategies tailored to variable latency.

\subsection{Impact of Rotation Noise} Cross-agent feature fusion relies on accurate relative poses between agents, where pose estimation errors can cause spatial misalignment and degrade cooperative perception performance. To investigate this effect, we follow V2X-ViT \cite{xu2022v2x}'s methodology by injecting noise into rotation matrices under two settings: (1) noise applied solely to the cross agent alignment module's inputs, and (2) global noise applied throughout the framework. 
As illustrated in \cref{fig:sup_robust}(c), global noise causes significant performance degradation, while introducing noise solely to the alignment module results in marginal decline. This reveals that although the alignment module takes pose parameters as input, it learns additional implicit information from features during training to achieve robust multi-agent feature alignment in latent space. The observed system-level sensitivity primarily stems from reference point perturbations rather than the alignment mechanism itself. Reducing the impact of pose noise remains an important direction for future research.

\begin{figure*}[!htbp]
    \centering
    
    \begin{subfigure}[b]{0.45\textwidth}
        \centering
        \includegraphics[width=\textwidth]{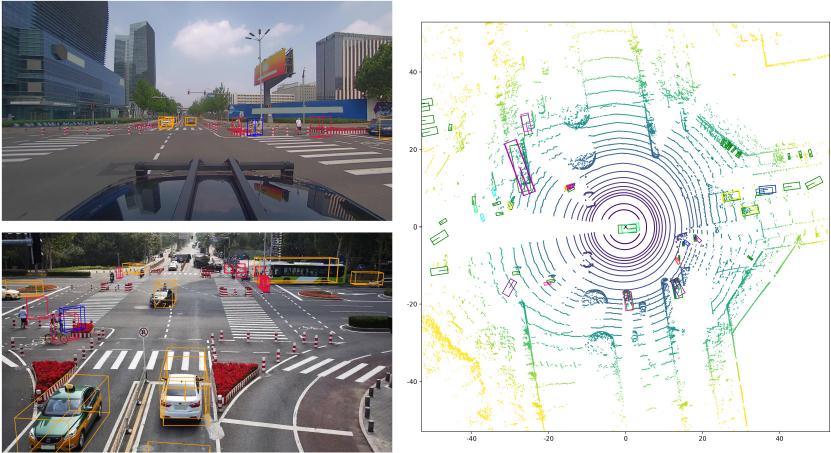}
        \caption{Example Sequence 1 - 1}
    \end{subfigure}
    \hspace{1cm}
    \begin{subfigure}[b]{0.45\textwidth}
        \centering
        \includegraphics[width=\textwidth]{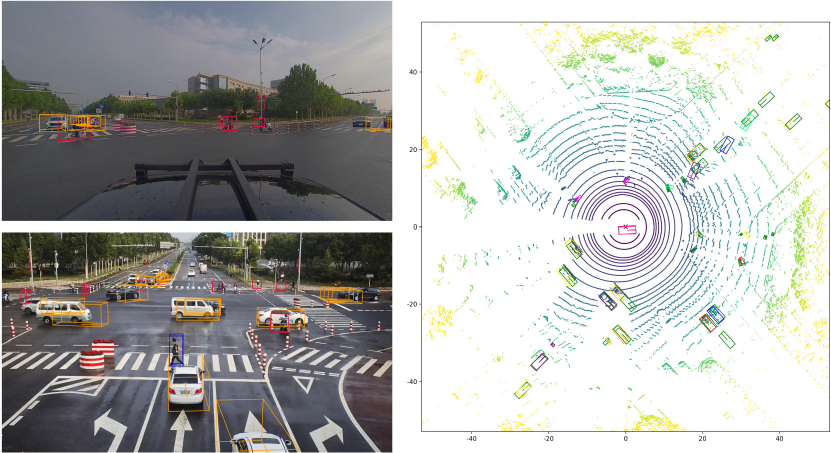}
        \caption{Example Sequence 2 - 1}
    \end{subfigure}
    \vspace{0.3cm}

    \begin{subfigure}[b]{0.45\textwidth}
        \centering
        \includegraphics[width=\textwidth]{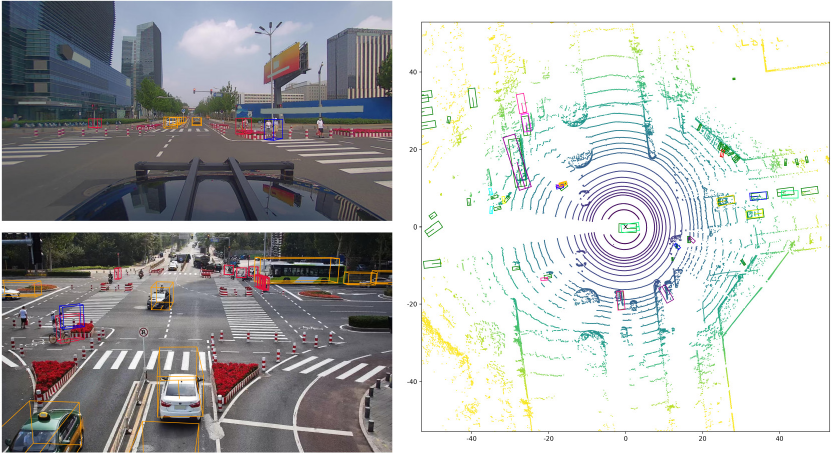}
        \caption{Example Sequence 1 - 2}
    \end{subfigure}
    \hspace{1cm}
    \begin{subfigure}[b]{0.45\textwidth}
        \centering
        \includegraphics[width=\textwidth]{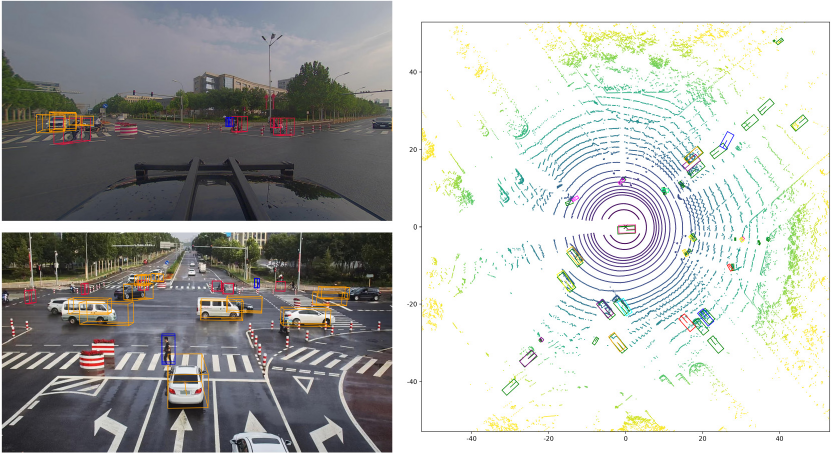}
        \caption{Example Sequence 2 - 2}
    \end{subfigure}
    \vspace{0.3cm}

    \begin{subfigure}[b]{0.45\textwidth}
        \centering
        \includegraphics[width=\textwidth]{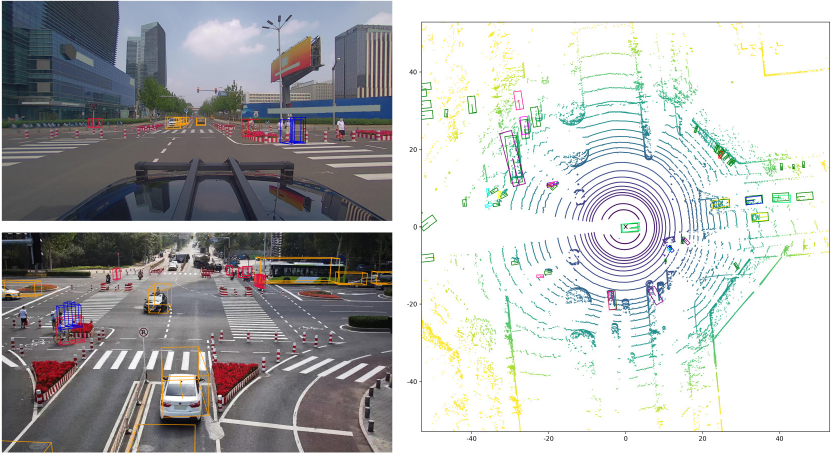}
        \caption{Example Sequence 1 - 3}
    \end{subfigure}
    \hspace{1cm}
    \begin{subfigure}[b]{0.45\textwidth}
        \centering
        \includegraphics[width=\textwidth]{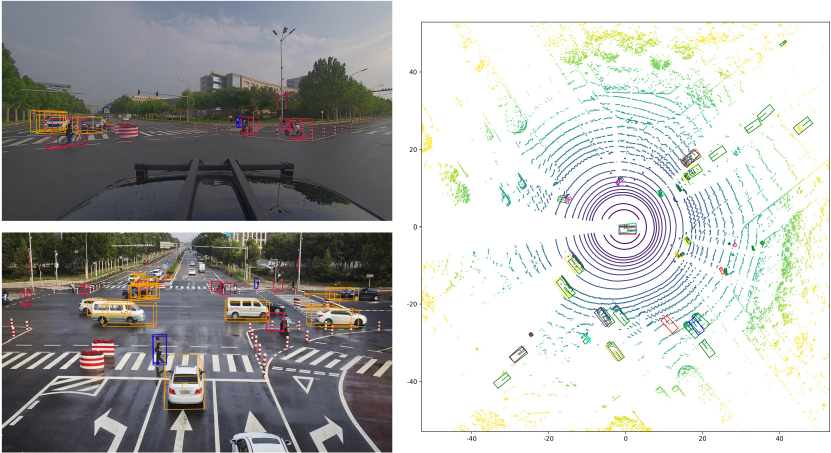}
        \caption{Example Sequence 2 - 3}
    \end{subfigure}
    \vspace{0.3cm}

    \begin{subfigure}[b]{0.45\textwidth}
        \centering
        \includegraphics[width=\textwidth]{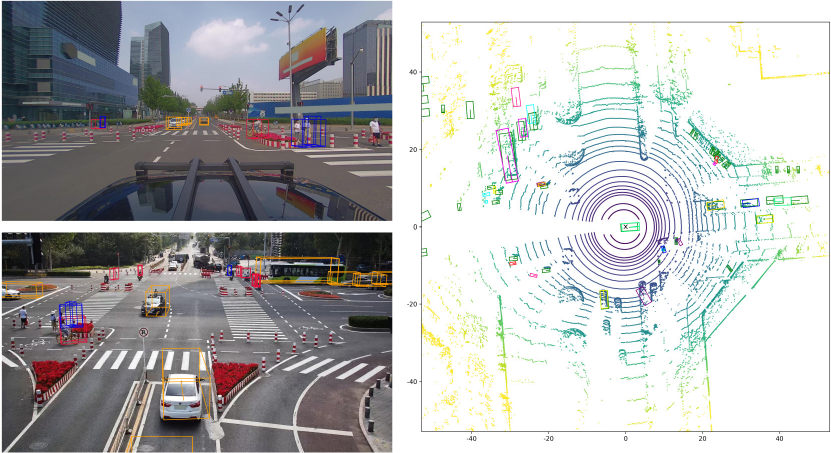}
        \caption{Example Sequence 1 - 4}
    \end{subfigure}
    \hspace{1cm}
    \begin{subfigure}[b]{0.45\textwidth}
        \centering
        \includegraphics[width=\textwidth]{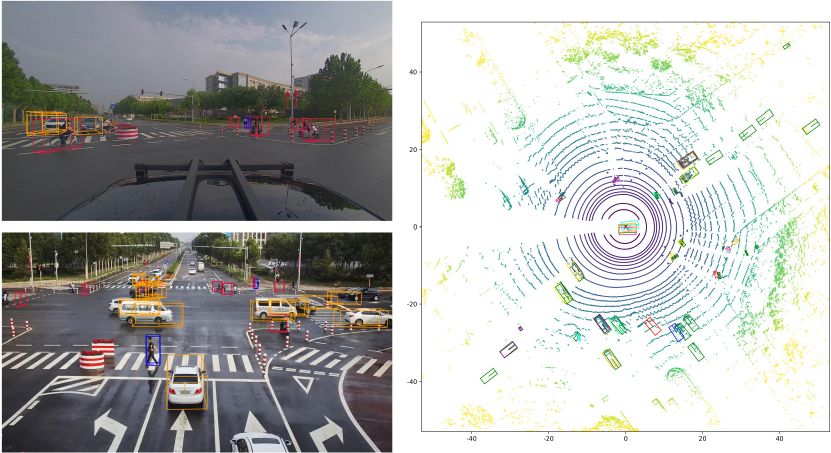}
        \caption{Example Sequence 2 - 4}
    \end{subfigure}

    \caption{\textbf{Visualization of CoopTrack's tracking results on the V2X-Seq Dataset.} The two columns display two different sequences from the validation split of the dataset, with four rows representing consecutive time steps. Each subfigure is divided into three parts: the top-left shows results from the vehicle perspective, the bottom-left shows results from the roadside perspective, and the right side presents the Bird's Eye View (BEV) visualization. In the forward-looking perspective, 3D bounding boxes are color-coded by object category: orange for vehicles, red for cyclists, and blue for pedestrians. In the BEV, the LiDAR point cloud is visualized for better presentation. For bounding boxes, we use green to represent the ground truth, while the colors of tracking results are randomly selected from a pool of colors based on their IDs, ensuring that each object maintains a consistent color over time.}
    \label{fig:result_visualization}
\end{figure*}

\section{Qualitative Analyses}
\label{supp:visualization}

\subsection{Association Results}
As shown in \cref{fig:sup_association_visualization}, we visualize instance association results for specific frames in a sequence within the ego-vehicle-centric BEV coordinate frame, where each instance is represented by its reference point. Successfully matched instance pairs, including both vehicle and infrastructure instances, are assigned unique IDs and labeled with colors from a predefined palette based on their IDs. In contrast, unmatched vehicle instances and roadside instances are displayed in blue and green, respectively. Note that due to the close proximity of instances, there is overlap in the figure. These results demonstrate that our association approach achieves pairing without relying solely on Euclidean distance, exhibiting strong robustness.
For example, consider a pair of instances located within the 20–30 meter range along the x-axis and near -10 meters along the y-axis (pair-11 in \cref{fig:sup_association_visualization}(a), pair-14 in \cref{fig:sup_association_visualization}(b), pair-12 in \cref{fig:sup_association_visualization}(c), and pair-13 in \cref{fig:sup_association_visualization}(d)). Due to inevitable inaccuracies in the reference points of vehicle and roadside instances, their relative distance varies and reaches a maximum in \cref{fig:sup_association_visualization}(c) yet stable association is maintained. This is because our association module comprehensively considers multi-dimensional instance features and relative positional relationships, thereby providing more reliable information for downstream aggregation module.

\subsection{Tracking Results}
As illustrated in \cref{fig:result_visualization}, we visualize the tracking results of our proposed CoopTrack on two representative sequences from V2X-Seq \cite{yu2023v2x}, aiming to intuitively demonstrate the superiority of our approach.
Each subfigure comprises three components: the vehicle-side input image positioned at the top-left, the roadside input image at the bottom-left, and the tracking results in the BEV view on the right. In the images, colors correspond to object categories: orange denotes vehicles, red denotes cyclists, and blue denotes pedestrians. In the BEV view, green bounding boxes represent ground truth, while colored boxes show tracking results with colors assigned by IDs to demonstrate temporal consistency.
In the two sequences provided, we can observe that, thanks to the cooperative information from the roadside, the ego-vehicle can continuously track instances behind and to the sides of it, achieving comprehensive perception results. This highlights the significant advantage of cooperative perception over single-vehicle perception. Furthermore, despite relying solely on images and lacking depth information input, CoopTrack has also achieved relatively precise localization, demonstrating the accurate tracking capability of our method in complex traffic scenarios.

\end{document}